\documentclass[10pt,twocolumn,letterpaper]{article}

\usepackage[pagenumbers]{cvpr}      

\usepackage{graphicx}
\usepackage{amsmath}
\usepackage{amssymb}
\usepackage{booktabs}
\usepackage{multirow}
\usepackage{makecell}

\usepackage[pagebackref,breaklinks,colorlinks]{hyperref}

\usepackage[capitalize]{cleveref}
\crefname{section}{Sec.}{Secs.}
\Crefname{section}{Section}{Sections}
\Crefname{table}{Table}{Tables}
\crefname{table}{Tab.}{Tabs.}

\def\cvprPaperID{3486} 
\def\confName{CVPR}
\def\confYear{2022}

\begin{document}

\title{PatchMVSNet: Patch-wise Unsupervised Multi-View Stereo for Weakly-Textured Surface Reconstruction}

\author{Haonan Dong \hspace{2em}Jian Yao$^{*}$\\
School of Remote Sensing and Information Engineering, Wuhan University\\
{\tt\small haonandong@whu.edu.cn}, {\tt\small jian.yao@whu.edu.cn}
}

\maketitle

\begin{abstract}
  Learning-based multi-view stereo (MVS) has gained fine reconstructions on popular datasets. However, \mbox{supervised} learning methods require ground truth for training, which is hard to be collected, especially for the large-scale datasets. Though nowadays unsupervised learning methods have been proposed and have gotten gratifying results, those methods still fail to reconstruct intact results in challenging scenes, such as weakly-textured surfaces, as those methods primarily depend on pixel-wise photometric consistency which is subjected to various illuminations. To alleviate matching ambiguity in those challenging scenes, this paper proposes robust loss functions leveraging constraints beneath multi-view images: 1) Patch-wise photometric consistency loss, which expands the receptive field of the features in multi-view similarity measuring, 2) Robust two-view geometric consistency, which includes a cross-view depth consistency checking with the minimum occlusion. Our unsupervised strategy can be implemented with arbitrary depth estimation frameworks and can be trained with arbitrary large-scale MVS datasets. Experiments show that our method can decrease the matching ambiguity and particularly improve the completeness of weakly-textured reconstruction. Moreover, our method reaches the performance of the state-of-the-art methods on popular benchmarks, like DTU, Tanks and Temples and ETH3D. The code will be released soon.
\end{abstract}


\section{Introduction}
\label{sec:intro}
Dense reconstruction aims to reconstruct the three-dimension (3D) dense structure from a collection of images and is widely applied in augmented reality (AR)\cite{azuma1997survey}, robotics and modern urban planning. As the core technology of dense reconstruction, multi-view stereo (MVS) can recover the 3D point for each target pixel in images, which uses known poses of images and the sparse point cloud from structure from motion (SfM). The mainstream methods for MVS can be classified as: direct point generation\cite{1388267}, voxels\cite{101145,4270218}, and depth fusion\cite{choe2021volumefusion,xu2020planar}. Among these methods, depth fusion is the most popular way as it can estimate the depth map of the reference image by multi-view geometry and then fuse all the depth maps into a dense point cloud without much difficulty.

\begin{figure}[t]
  \centering
  \includegraphics[width=\linewidth]{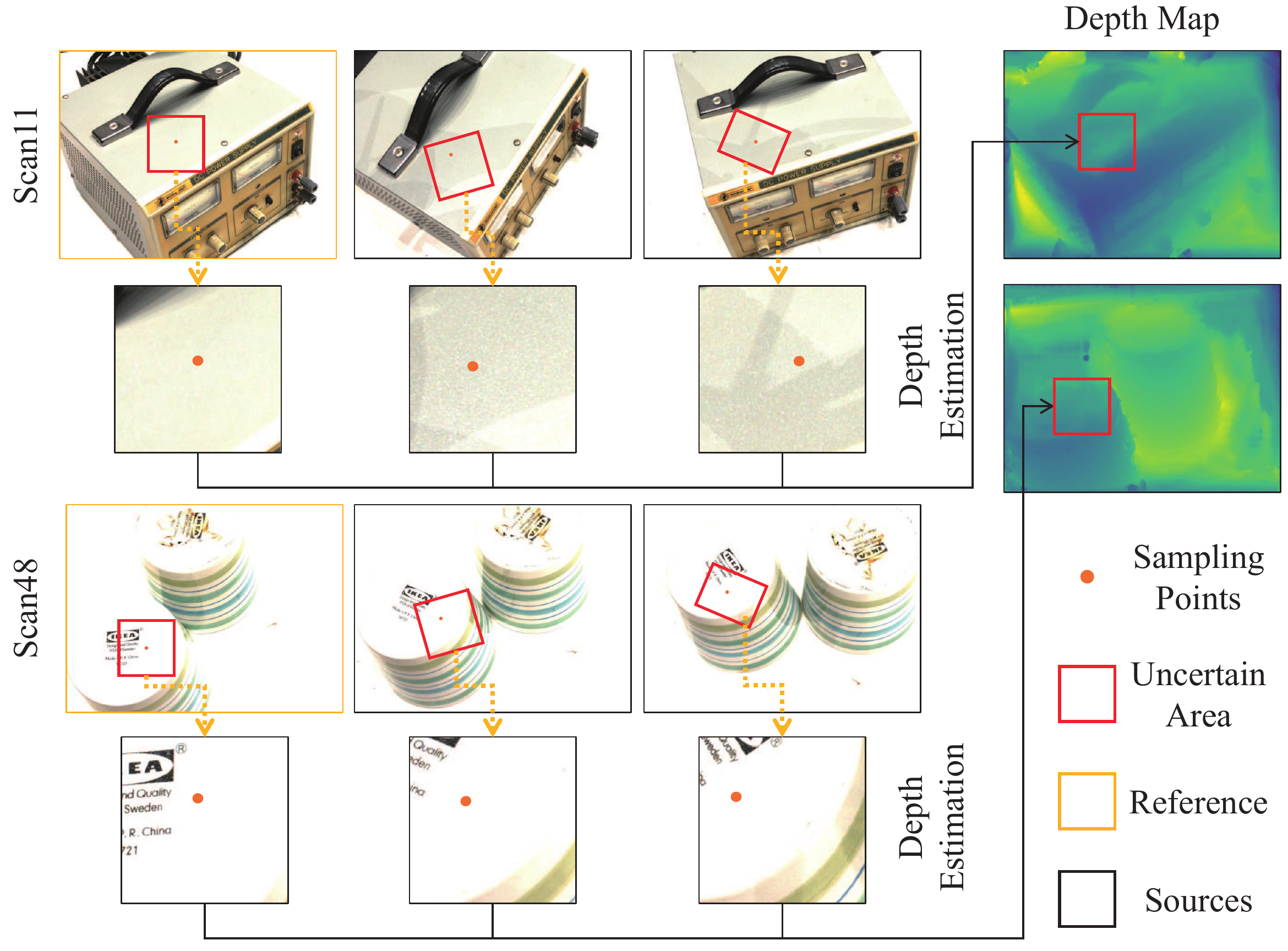}\\[-0.75em]
  \caption{Drawback of the pixel-wise photometric consistency. Various brightness and weak texture result in ambiguous matchings, where those matchings are highly uncertain for depth estimation. The upper object represents error matchings from various brightness, and the lower object represents ambiguity matchings from weakly-textured surfaces.}
  \label{fig:drawback}
  \vspace{-1.5em}
\end{figure}

To acquire the depth map of the images, traditional methods for depth fusion initially use the plane-sweeping algorithm\cite{collins1996space} to compute the disparity information from the stereo pairs, which construct a cost volume in disparity space to infer a suitable disparity of each pixel by maximizing the photometric consistency of the correspondences. Then two-view stereo is extended to multi-view stereo by calculating a redundant photometric consistency of multiple images to get the best depth map of the reference image\cite{gallup2007real,galliani2015massively}. After getting the depth maps, traditional methods fuse all depth maps with the geometric checking and get the point clouds. However, there are two basic problems for the traditional plane-sweeping technique: 1) the region of interest has a specific depth range\cite{bleyer2011patchmatch}, which limits the scope of the reconstruction, 2) it is difficult to obtain certain matchings for weakly-textured surfaces, as illustrated in \mbox{Figure~\ref{fig:drawback}}. These two problems impair the accuracy and the completeness of traditional methods, especially for the reconstruction of high-resolution images.

With the development of deep learning, convolutional neural networks (CNNs) are also applied into depth-fusion-based MVS methods, named learning-based MVS methods. Whether to train the CNNs with ground truth or not, learning-based MVS methods can be further categorized into supervised methods and unsupervised methods. Supervised methods have provided useful solutions to solve the problems in traditional methods. For the first problem, those methods can not only estimate a larger range of depth, but sample more precise depth intervals, with which they reconstruct a more complete and accurate result\cite{yao2018mvsnet,gu2020cascade}. For the second obstacle, supervised methods can train CNNs to extract more robust features to obtain precise matchings with ground truth and obtain intact results even in weakly-textured scenes. However, one limitation of supervised methods is that those methods require ground truth, which costs much to be collected. Lacking training data makes the performance of the supervised methods worse than the traditional methods\cite{schoenberger2016mvs,romanoni2019tapa,xu2020planar,xu2019multi} in the reconstruction of large-scale datasets\cite{schoeps2017cvpr}.

Unsupervised methods need no ground truth, but the issus is how to train the network? Fortunately, by digging out the principle of image rendering, it is possible to reconstruct the rendered reference image by sampling the pixel intensity from the source images with known poses and the depth map of the reference image\cite{chen2021mvsnerf,ruckert2021adop}, which enables unsupervised methods to be trained by leveraging the information of multi-view geometry instead of ground truth. For example, the pixel-wise photometric consistency\cite{khot2019learning,huang2021m3vsnet} and the cross-view depth consistency\cite{dai2019mvs2} have already been exploited to train CNNs and some unsupervised methods\cite{xu2021digging,Yang_2021_CVPR} even outperform supervised methods by the multi-stage processing strategy.
\begin{figure}[t]
  \centering
  \begin{subfigure}{0.26\textwidth}
    \centering
    \includegraphics[height=3.6cm]{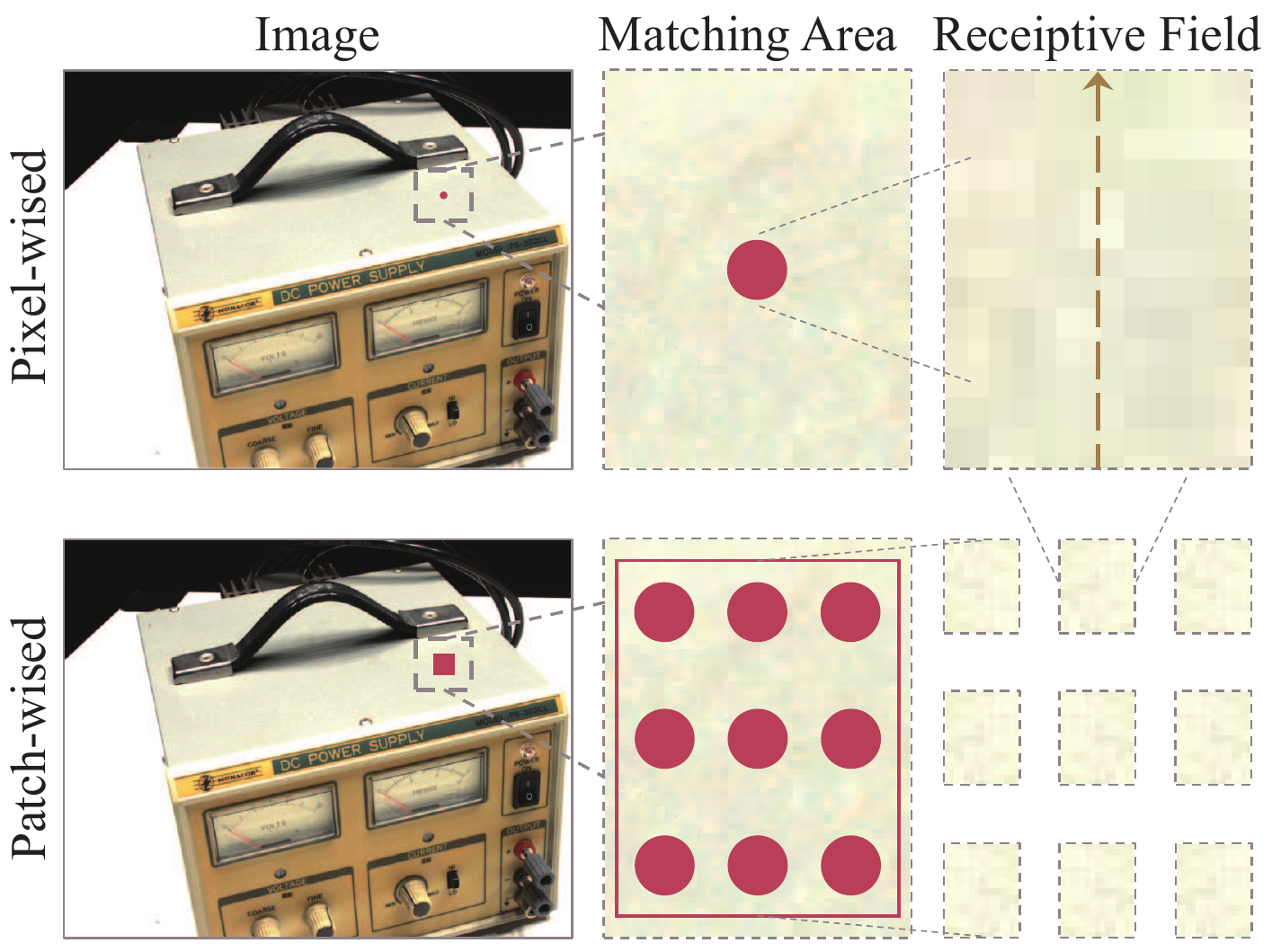}
    \caption{Receptive field comparison}
  \end{subfigure}
  \hfill
  \begin{subfigure}{0.2\textwidth}
    \centering
    \includegraphics[height=3.6cm]{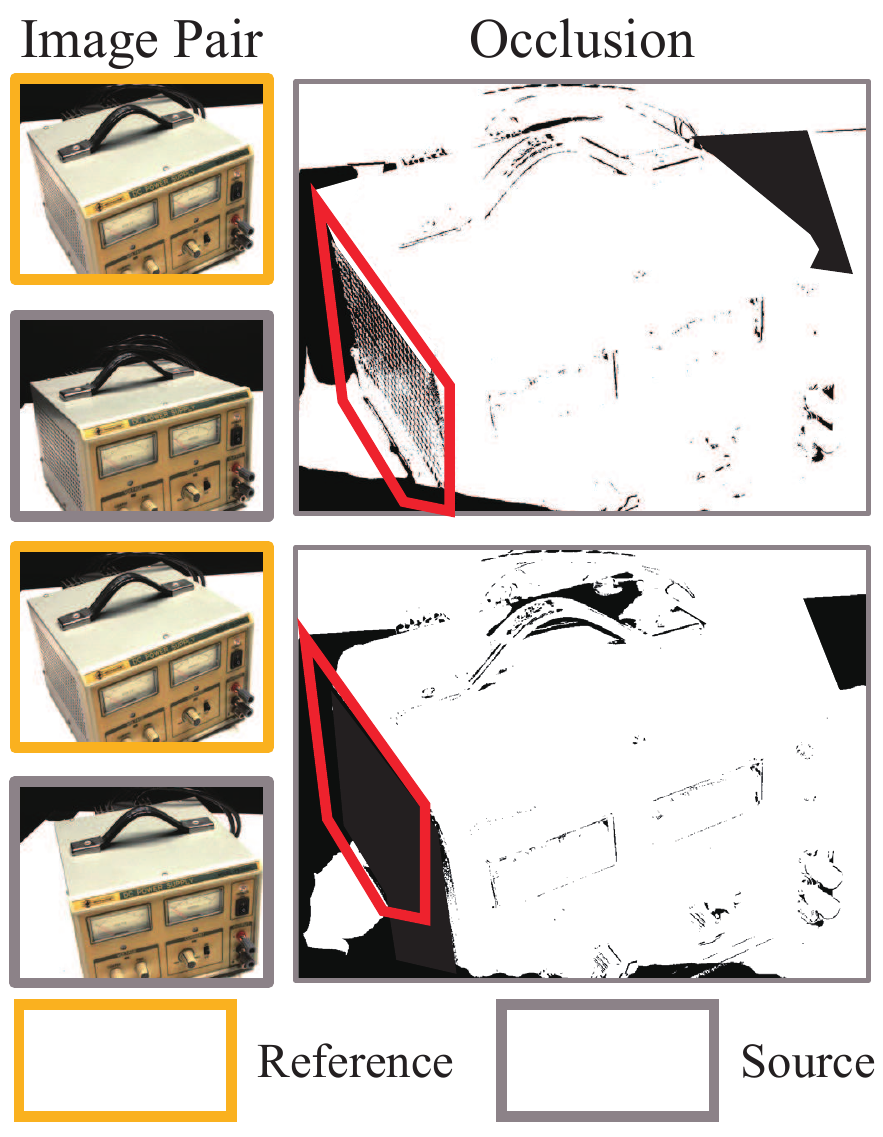}
    \caption{Occlusion}
  \end{subfigure}\\[-0.75em]
  \caption{Problems in current unsupervised methods: (a) receptive field comparison for a single pixel and a patch, (b) occlusion from view changing in depth-fusion-based methods, where black region means occlusion. The reference image in (b) is the 26$^{th}$ image of scan11 in DTU evaluation datasets. The upper image with the gray outline is the 1$^{st}$ source image and the lower image with the gray outline is the 3$^{rd}$ source image. As illustrated in the red polygons, the best source image have the minimum occlusion than other source images.}
  \label{fig:us-short}
  \vspace{-1.5em}
\end{figure}
\par
Nonetheless, the pixel-wise photometric consistency is handily subjected to challenging environments like weakly-textured surface, since it relies on the pixel-wise similarity. As illustrated in \mbox{Figure~\ref{fig:us-short}(a)}, the red pixel with the weak texture has a small receptive field and the change in intensity is slight, thus, it is hard to ensure whether the pixel in the reference image is certainly matched with the pixel from the rendered reference image. For this kind of pixels, the matching will cost much even with the right depth. Absent from ground truth, the pixel-wise photometric consistency is accepting mistaken matchings especially on weakly-textured surfaces. So training CNNs by the pixel-wise photometric consistency can only obtain intact results in edges and richly-textured surfaces\cite{xu2021self}. Moreover, the geometric consistency, used to further alleviate the matching ambiguity, is also restricted, since it often measures a cross-view depth consistency which should take occlusion into consideration. The occlusion information indicates if a pixel should be included in the depth consistency computation, which is shown in \mbox{Figure~\ref{fig:us-short}(b)}. If the pixel of the source image is not visible on the reference image, it will bring in ambiguity in similarity measuring. But it is hard to get reliable visibility information in unsupervised methods\cite{zhang2020visibility}.
\par
To these issues in unsupervised methods, in this paper, we present PatchMVSNet, which novelly introduces robust loss functions for unsupervised learning. In traditional methods\cite{xu2019multi,bleyer2011patchmatch}, a support domain or a patch and hand-crafted features are always used to measure the similarity of the matching correspondences, such as Normalized Cross Corelation (NCC) and Zero Normalized Cross Corelation (ZNCC). Naturally, the bigger receptive field is, the more context information and the more robust matchings can the network get, as shown in \mbox{Figure~\ref{fig:us-short}(a)}. So in \mbox{PatchMVSNet}, a patch is used to compute the robust patch-wise photometric consistency. Moreover, a robust geometric consistency loss, considering the occlusion, is used to decrease the ambiguity of the correspondences. The high-level features alignment module is also integrated to strengthen our method as it also alleviates the influence from brightness\cite{huang2021m3vsnet}.

The contributions of our work are summarized as follows:\\[-2em]
\begin{itemize}
  \item A patch-wise photometric consistency loss is introduced to decrease outliers and ambiguous matchings.\\[-2em]
  \item A robust geometric consistency loss is proposed in the asymmetric depth estimation structure.\\[-2em]
  \item Our method achieves a competitive performance to the state-of-the-art unsupervised methods, especially in the weakly-textured datasets, such as \textit{ETH3D}.
\end{itemize}

\begin{figure*}
  \centering
  \includegraphics[width=\textwidth]{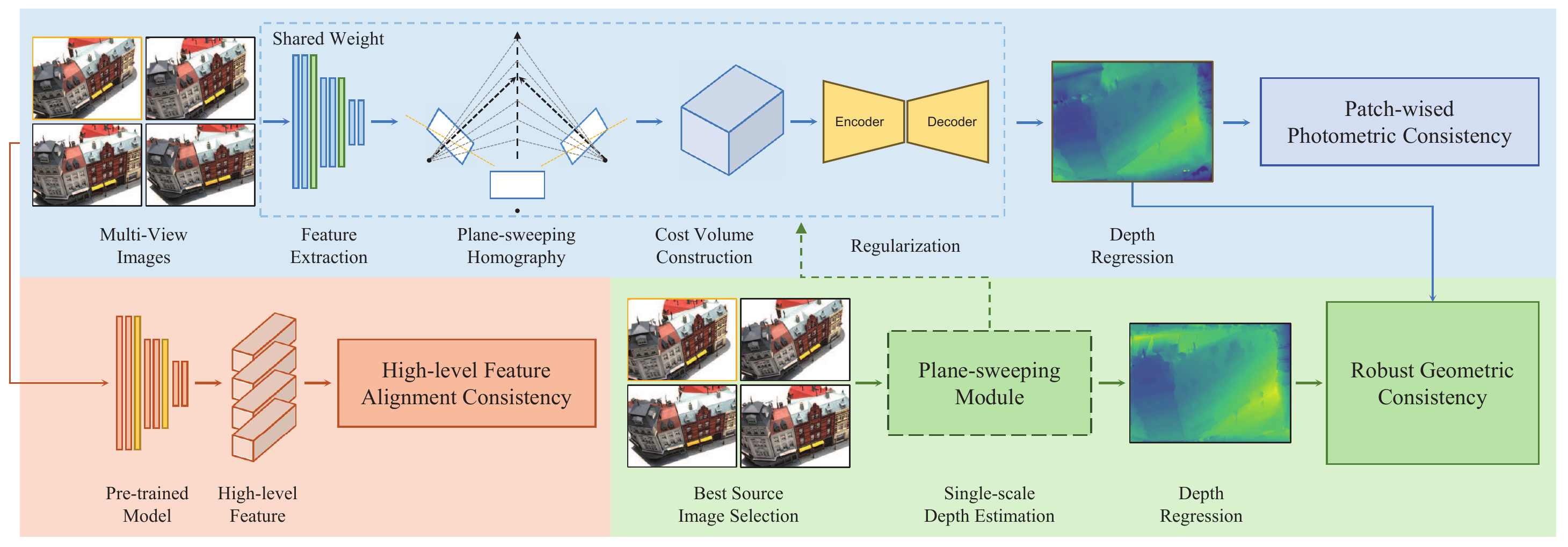}\\[-0.75em]
  \caption{Illustration of framework of our proposed PatchMVSNet. The blue part represents the basic backbone for depth estimation, regressing the reference's depth by the patch-wise photometric consistency (Section~\ref{sec:photo-loss}). The green part represents the geometric consistency module  (Section~\ref{sec:cross-view-loss}), which regresses the depth map of the best source(Section~\ref{sec:src-selection}). The orange part represents the high-level feature extraction and alignment module (Section~\ref{sec:high-level}). }
  \label{fig:loss-frame}
  \vspace{-1.5em}
\end{figure*}

\vspace{-1em}
\section{Related Work}
\label{sec:related}
\noindent
\textbf{Traditional MVS}. Prevailing traditional works mainly follows the pipeline of PatchMatchStereo\cite{bleyer2011patchmatch}, which initially samples a random depth for each pixel, then calculates the cost volume and propagates the inlier depth to pixels with lower confidences. Gipuma\cite{galliani2015massively} firstly introduced PatchMatchStereo into the multi-view scenario, utilizing a propagation strategy to exploit more reliable depth candidates for each pixel. To increase the efficiency of depth estimation, Sch{\"o}nberger \textit{et al.}\cite{schoenberger2016mvs} developed COLMAP, a complete pipeline including SfM, MVS and mesh reconstruction, where they utilized joint estimation of the depth and normal information to optimize the matching cost function. To improve the completeness of weakly-textured scenes, TAPA-MVS\cite{romanoni2019tapa} and ACMP\cite{xu2020planar} leveraged a prior information of the plane to reconstruct more intact results. Though the performance of traditional methods leads in high-resolution datasets\cite{schoeps2017cvpr}, the efficiency is still limited.
\par
\noindent
\textbf{Supervised MVS}. Yao \textit{et al.}\cite{yao2018mvsnet} presented MVSNet which primarily used deep learning framework to implement the plane-sweeping technique. To decrease the memory cost for high-resolution reconstruction, Gu \textit{et al.}\cite{gu2020cascade} extracted multi-scale features with the feature pyramid network (FPN)\cite{lin2017feature} and constructed the cascaded cost volume to reduce memory consumption of regularization. To enhance the matching certainty with larger receptive field, Luo \textit{et al.}\cite{luo2019p} utilized a patch-wise similarity confidence for cost volume construction. Until Wang \textit{et al.}\cite{wang2021patchmatchnet} implemented PatchMatchNet, most supervised methods\cite{yao2019recurrent, xue2019mvscrf,sormann2020bp,YU2021448} continue the principle of the plane-sweeping algorithm to improve the performance of networks. Although supervised methods have mitigated problems of traditional methods, lacking ground truth impairs the ability of generalization.
\par
\noindent
\textbf{Unsupervised MVS}. Khot \textit{et al.}\cite{khot2019learning} digged the robust pixel-wise photometric consistency to train the network and achieved a significant improvement in DTU datasets. To examine the effectiveness of geometric consistency, Dai \textit{et al.}\cite{dai2019mvs2} used a symmetric structure to additionally exploit the cross-view depth consistency. To further mitigate the influence of uncertain matching, Huang \textit{et al.}\cite{huang2021m3vsnet} extended Khot's work and extracted high-level features from a pre-trained model to improve the quality of the depth map. With a multi-stage strategy, latest works mainly paid attention to train the network with the assistants from third-parties, such as a rendered mesh\cite{Yang_2021_CVPR}, optical flow\cite{xu2021digging}. However, the accuracy and the completeness of reconstructions in weakly-textured scenes are still limited.

\vspace{-0.25em}
\section{Method}
\label{sec:method}
Our method will be introduced in two sections. The essential backbone of our method and the selection of the best source image are introduced in \mbox{Section~\ref{sec:backbone}}. Our robust loss functions are explained in \mbox{Section~\ref{sec:loss}}.
\subsection{Backbone}
\label{sec:backbone}
Two standard architectures, classified by Darmon \textit{et al.}\cite{darmon2021deep}, are selected as our backbones: MVSNet architecture and multi-scale architecture. For convenience, our method is illustrated with the MVSNet architecture, as shown in \mbox{Figure~\ref{fig:loss-frame}}. Deep features are firstly extracted by CNNs and aligned into the same coordinate by the plane-sweeping technique with the depth map of the reference image. Then a cost volume is constructed by the variance metric and is regularized to compute the depth probabilities by 3D U-Net\cite{ronneberger2015u}. Finally, the depth  $d{\boldsymbol{(p)}}$ of the pixel $\boldsymbol{\mathrm{p}} = (x,y)^\mathrm{T}$ can be regressed by computing the expectation along the depth direction:\\[-0.25em]
\begin{equation}
  d{\boldsymbol{(p)}}  = \sum\nolimits_{d = D_\mathrm{min}}^{D_\mathrm{max}} d \cdot P(\boldsymbol{\mathrm{p}},d),
  \label{eq:depth_regress}
\end{equation}\\[-0.5em]
where $D_\mathrm{min},D_\mathrm{max}$ mean the minimum and maximum values of the depth range, and $P(\boldsymbol{\mathrm{p}},d)$ means the probability of taking the depth $d$ at the pixel $\boldsymbol{\mathrm{p}}$.

\vspace{-1em}
\subsubsection{Best Source Image Selection}
\label{sec:src-selection}
Geometric consistency is an important constraint in unsupervised methods, since it can evaluate the quality of the depth map with the redundant cross-view information. Most learning-based MVS methods are the asymmetric structure, which means they only infer the depth map of the reference image each time. Though the symmetric structural method\cite{dai2019mvs2} can infer multiple depth maps every time, it is extremely memory-consuming. Thus, we still use the asymmetric structure to estimate multiple depth maps. Moreover, according to the algorithm\cite{schoenberger2016mvs} of view selection for the asymmetric structure, the reference image will take images as sources that include enough overlap with it. Images with more co-visible 3D sparse points with the reference image will get higher scores and the algorithm always selects the top $N$ images as sources of the reference image. Considering the influence of occlusion and the memory limitation, only the view with the highest score is selected to build a minimum-occlusion pair with the reference image for checking the geometric consistency. So each time two depth maps of the reference image and of the best source image are estimated.
\par
\noindent
Implemented with the multi-scale architecture\cite{gu2020cascade}, our method should sample new depth hypotheses of each image from the depth map of the former stage, since the plane-sweeping technique should build fronto-parallel planes with different depth hypotheses when the reference image changes. Thus, in each stage $k$, for each pixel $\boldsymbol{\mathrm{p}}$, the depth sampling range $\mathcal{R}_{k}\boldsymbol{\mathrm{(p)}}$ is induced as $\mathcal{R}_{k}\boldsymbol{\mathrm{(p)}} = \{d_{k-1}{\boldsymbol{\mathrm{(p)}}} - {\Delta d_{k}}, d_{k-1}{\boldsymbol{\mathrm{(p)}}} + {\Delta d_{k}}\}$, where $d_{k}\boldsymbol{\mathrm{(p)}}$ is the depth of $\boldsymbol{\mathrm{p}}$ in the stage $k$. $\Delta d_{k} = n_{k}\cdot \gamma_{k}$ is the depth sampling range computed by the depth hypotheses numbers $n_{k}$ and the depth interval $\gamma_{k}$.

\subsection{Loss}
\label{sec:loss}
\subsubsection{Photometric Consistency}
\label{sec:photo-loss}
\textbf{Pixel-wise Photometric Consistency}. Photometric consistency measures the similarity of pixels that represent the same 3D point $\boldsymbol{\mathrm{P}} = (X,Y,Z)^\mathrm{T}$. Given a reference image $\boldsymbol{\mathrm{I}}_\mathrm{ref}$ and selected source images $\{\boldsymbol{\mathrm{I}}_\mathrm{src}^{(i)}\}_{i=1}^N$, the 3D point $\boldsymbol{\mathrm{P}}$ can be projected into the image plane as the pixel $\boldsymbol{\mathrm{p}}$. The intensity similarity in multi-view images is always computed by warping the pixel $\boldsymbol{\mathrm{p}}_\mathrm{ref} \in \boldsymbol{\mathrm{I}}_\mathrm{ref}$ into $\boldsymbol{\mathrm{I}}_\mathrm{src}^{(i)}$ by:\\[-0.25em]
\begin{equation}
  \hat{\boldsymbol{\mathrm{p}}}_\mathrm{src}^{(i)} = \boldsymbol{\mathrm{K}}_\mathrm{ref}\boldsymbol{\mathrm{T}}_\mathrm{ref}\boldsymbol{\mathrm{T}}_{i}^{-1}\boldsymbol{\mathrm{K}}_{i}^{-1}(d{(\boldsymbol{\mathrm{p}}_\mathrm{ref})} \cdot \boldsymbol{\mathrm{p}}_\mathrm{ref}),
  \label{eq:warp}
\end{equation}\\[-0.25em]
where $\hat{\boldsymbol{\mathrm{p}}}_\mathrm{src}^{(i)} \in \boldsymbol{\mathrm{I}}_\mathrm{src}^{(i)}$ means the warped pixel from $\boldsymbol{\mathrm{I}}_\mathrm{ref}$ to $\boldsymbol{\mathrm{I}}_\mathrm{ref}^{(i)}$; $\boldsymbol{\mathrm{K}}_{i}, \boldsymbol{\mathrm{T}}_{i}$ means the intrinsic and extrinsic matrixes of $\boldsymbol{\mathrm{I}}_\mathrm{ref}^{(i)}$ and $\boldsymbol{\mathrm{K}}_\mathrm{ref}, \boldsymbol{\mathrm{T}}_\mathrm{ref}$ means the intrinsic and extrinsic matrixes of $\boldsymbol{\mathrm{I}}_\mathrm{ref}$.
\par
\noindent
Then a rendered reference image $\hat{\boldsymbol{\mathrm{I}}}_\mathrm{src}^{(i)}$ with the same pixel coordinate of $\boldsymbol{\mathrm{I}}_\mathrm{ref}$ can be sampled from $\boldsymbol{\mathrm{I}}_\mathrm{src}^{(i)}$ by:\\[-0.25em]
\begin{equation}
  \hat{\boldsymbol{\mathrm{I}}}_\mathrm{src}^{(i)}(\boldsymbol{\mathrm{p}}_\mathrm{ref}) = \boldsymbol{\mathrm{I}}_\mathrm{src}^{(i)}(\hat{\boldsymbol{\mathrm{p}}}_\mathrm{src}^{(i)}).
  \label{eq:pixel_sample}
\end{equation}\\[-0.25em]
Previous works\cite{khot2019learning,xu2021self} utilized the pixel-wise photometric consistency loss by measuring the absolute intensity difference between $\boldsymbol{\mathrm{I}}_\mathrm{ref}$ and $\hat{\boldsymbol{\mathrm{I}}}_\mathrm{src}^{(i)}$:\\[-0.25em]
\begin{equation}
  \boldsymbol{\mathcal{L}}_\mathrm{photo}=\frac{1}{N}\sum\nolimits^{N}_{i=1}|\boldsymbol{\mathrm{I}}_\mathrm{ref} - \hat{\boldsymbol{\mathrm{I}}}_\mathrm{src}^{(i)}|\odot \boldsymbol{\mathrm{M}}_\mathrm{ref},
  \label{eq:photometric_pixel}
\end{equation}\\[-0.25em]
where $\boldsymbol{\mathrm{M}}_\mathrm{ref}$ means the mask of valid pixels within the depth range and $\odot$ means the element-wise product.
\par
\noindent
A single point has the limited receptive field (see \mbox{Figure~\ref{fig:us-short}(a)}) and is not reliable to estimate depth on weakly-textured surfaces. But a patch, including more global information, will make the center pixel more distinguished, so the pixel-wise photometric consistency is extended into the patch-wise photometric consistency in order to increase the matching certainty.
\par
\noindent
\textbf{Patch-wise Photometric Consistency}. We define a $m^2$-sized square patch centering on the pixel $\boldsymbol{\mathrm{p}}$ as $\boldsymbol{\Omega}({\boldsymbol{\mathrm{p}}})$. The patch $\boldsymbol{\Omega}({\boldsymbol{\mathrm{p}}})$ is initially reshaped into a single-dimension vector $\boldsymbol{\omega}({\boldsymbol{\mathrm{p}}})$ and all vectors in each image are concatenated by the row-major order. The high-dimension ``images" $\{\boldsymbol{\mathcal{I}}_{\mathrm{src}}^{(i)}\}_{i=1}^N, \boldsymbol{\mathcal{I}}_{\mathrm{ref}}$ are then generated from the three-channel images $\{\boldsymbol{\mathrm{I}}_{\mathrm{src}}^{(i)}\}_{i=1}^N, \boldsymbol{\mathrm{I}}_{\mathrm{ref}}$. The shape of $\boldsymbol{\mathcal{I}}_{\mathrm{src}}^{(i)}$ is $C\times H\times W$, where $H, W$ are the height and the width of $\boldsymbol{\mathrm{I}}_{\mathrm{src}}^{(i)}$ and $C$ means the dimension number of $\boldsymbol{\omega}({\boldsymbol{\mathrm{p}}})$. The local patch is small so it can be treated as a plane\cite{yu2020p}, which assumes the patch shares the same depth of the center pixel. So like \mbox{Eq.~\ref{eq:warp}}, the patch-wise warping equation can be written as:\\[-0.25em]
\begin{equation}
  \hat{\boldsymbol{\omega}}({\boldsymbol{\mathrm{p}}}_\mathrm{src}^{(i)}) = \boldsymbol{\mathrm{K}}_\mathrm{ref}\boldsymbol{\mathrm{T}}_\mathrm{ref}\boldsymbol{\mathrm{T}}_{i}^{-1}\boldsymbol{\mathrm{K}}_{i}^{-1}(d({{\boldsymbol{\mathrm{p}}}_\mathrm{ref})} \cdot \boldsymbol{\omega}({\boldsymbol{\mathrm{p}}_\mathrm{ref}})),
  \label{eq:warp_patch}
\end{equation}\\[-0.25em]
where $\hat{\boldsymbol{\omega}}({\boldsymbol{\mathrm{p}}}_\mathrm{src}^{(i)}) \in \boldsymbol{\mathcal{I}}_{\mathrm{src}}^{(i)}$ means the warped patch from $\boldsymbol{\omega}({\boldsymbol{\mathrm{p}}_\mathrm{ref}}) \in \boldsymbol{\mathcal{I}}_\mathrm{ref}$ to $\boldsymbol{\mathcal{I}}_{\mathrm{ref}}^{(i)}$. As \mbox{Eq.~\ref{eq:pixel_sample}}, the high-dimension rendered reference image $\hat{\boldsymbol{\mathcal{I}}}_\mathrm{src}^{(i)}$ is reconstructed by:\\[-0.25em]
\begin{equation}
  \hat{\boldsymbol{\mathcal{I}}}_\mathrm{src}^{(i)}(\boldsymbol{\omega}({\boldsymbol{\mathrm{p}}_\mathrm{ref}})) = \boldsymbol{\mathcal{I}}_{\mathrm{src}}^{(i)}(\hat{\boldsymbol{\omega}}({\boldsymbol{\mathrm{p}}}_\mathrm{src}^{(i)})).
  \label{eq:patch_sample}
\end{equation}\\[-0.25em]
Finally, the patch-wise photometric consistency loss is formulated as:\\[-0.25em]
\begin{equation}
  \boldsymbol{\mathcal{L}}_\mathrm{patch}=\frac{1}{N}\sum\nolimits_{i=1}^{N}|\boldsymbol{\mathcal{I}}_\mathrm{ref} - \hat{\boldsymbol{\mathcal{I}}}_\mathrm{src}^{(i)}|\odot \mathcal{M}_\mathrm{ref},
  \label{eq:photometric_patch}
\end{equation}\\[-0.25em]
where $\mathcal{M}_\mathrm{ref}$ means the mask processed for patch-wise computation.
\begin{figure}[t]
  \centering
  \includegraphics[width=0.8\linewidth]{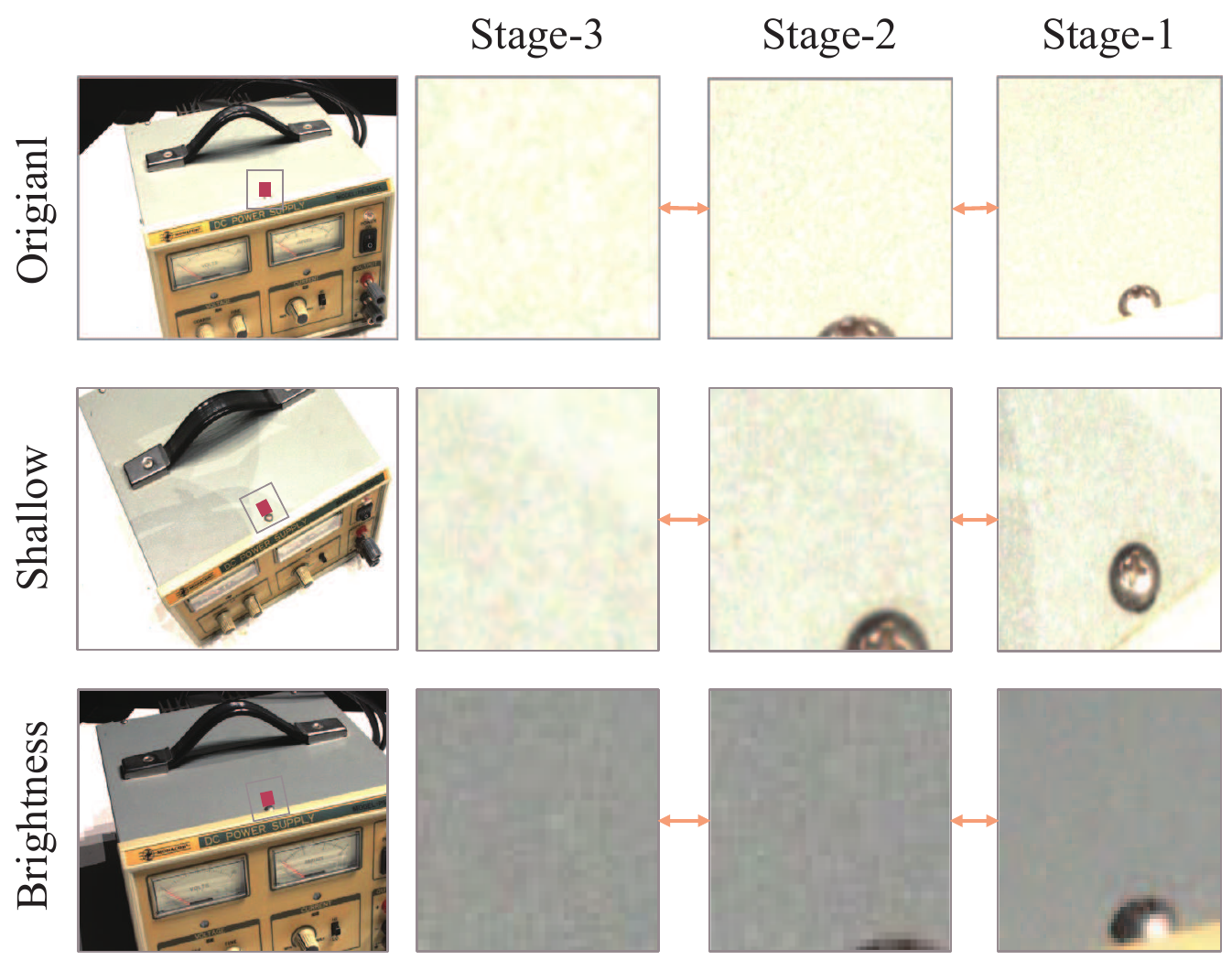}\\[-1em]
  \caption{Multi-scale patch merging strategy. Only the patches of the neighbor stages are merged with the red linking arrows. With patches of the same size, the sampling area becomes more noticeable as the resolution decreases (from left to right), even the source images is not identical to the reference image due to the shallows and the various brightness. }
  \label{fig:multi-scale-patch}
  \vspace{-1.5em}
\end{figure}
\par
\noindent
\textbf{Multi-Scale Patch Merging Strategy}. Similar to Needle-match\cite{lotan2016needle}, the multi-scale patches can be merged as a more discriminated descriptor for the target pixels in our multi-scale implementation. In the stage $k$, the patch of the target pixel $\boldsymbol{\mathrm{p}}$ is defined as $\boldsymbol{\Omega}_{k}{\boldsymbol{\mathrm{(p)}}}$. As the resolution decreases, features are more discriminative within the same size of the patch, as illustrated in \mbox{Figure~\ref{fig:multi-scale-patch}}. But the larger patch causes the loss of accuracy. Therefore we set the same size of patches for each stage, which means more context information can be obtained in the low resolution and more details can be preserved in the high resolution. Meanwhile, to follow the assumption that the patch shares the same depth, only the patches of two neighbor stages are merged by bilinear interpolation.
\par
\noindent
\textbf{Rendered Image Similarity Constraint}. Identical to most unsupervised methods, the structure similarity (SSIM)\cite{wang2004image} is used for examining the quality of $\hat{\boldsymbol{\mathcal{I}}}_\mathrm{src}^{(i)}$:\\[-0.25em]
\begin{equation}
  \boldsymbol{\mathcal{L}}_\mathrm{SSIM} = \frac{1}{N}\sum\nolimits_{i=1}^{N}\mathrm{SSIM}<\boldsymbol{\mathcal{I}}_\mathrm{ref},\hat{\boldsymbol{\mathcal{I}}}_\mathrm{src}^{(i)}> \odot \mathcal{M}_\mathrm{ref}.
  \label{eq:structure_similarity}
\end{equation}
\vspace{-1em}

\subsubsection{Geometric Consistency}
\label{sec:cross-view-loss}
\textbf{Robust Cross-View Geometric Consistency}. Apart from the photometric consistency, the geometric consistency is exploited to further diminish the matching ambiguity. After getting the minimum-occlusion pair and their depth maps, $\boldsymbol{\mathrm{I}}_\mathrm{ref}$ is warped into the coordinate of $\boldsymbol{\mathrm{I}}_\mathrm{src}^{(i)}$ with the depth map of $\boldsymbol{\mathrm{I}}_\mathrm{src}^{(i)}$ and the rendered image from the source image to the reference $\hat{\boldsymbol{\mathrm{I}}}_\mathrm{ref\rightarrow src}^{(i)}$ is obtained. Then $\hat{\boldsymbol{\mathrm{I}}}_\mathrm{ref\rightarrow src}^{(i)}$ is re-projected back to the coordinate of $\boldsymbol{\mathrm{I}}_\mathrm{ref}$ with the depth map of $\boldsymbol{\mathrm{I}}_\mathrm{ref}$. Then the cross-rendered reference image $\hat{\boldsymbol{\mathrm{I}}}_\mathrm{ref\rightleftarrows src}^{(i)}$ can be gotten by bilinear interpolation. After that a cross-view consistency checking can be made by measuring the similarity between $\hat{\boldsymbol{\mathrm{I}}}_\mathrm{ref\rightleftarrows src}^{(i)}$ and $\boldsymbol{\mathrm{I}}_\mathrm{ref}$. Since $\hat{\boldsymbol{\mathrm{I}}}_\mathrm{ref\rightleftarrows src}^{(i)}$ should be identical to $\boldsymbol{\mathrm{I}}_\mathrm{ref}$ from pixel to pixel. The robust geometric consistency loss is formulated as:\\[-0.25em]
\begin{equation}
  \boldsymbol{\mathcal{L}}_\mathrm{geometric}=\frac{1}{N}\sum\nolimits_{i=1}^{N}|\boldsymbol{\mathrm{I}}_\mathrm{ref} - \hat{\boldsymbol{\mathrm{I}}}_\mathrm{ref\rightleftarrows src}^{(i)}|\odot\boldsymbol{\mathrm{M}}_\mathrm{ref}.
  \label{eq:geometric_consistency}
\end{equation}\\[-0.25em]
\textbf{Depth Smooth Constraint}. As the patch will diminish the difference of intensities in edges, we bring in a depth smooth constraint\cite{mahjourian2018unsupervised} to regularize the depth estimation, which allows the depth to vary in the discontinued area and to keep the depth continued in the smooth surfaces:\\[-0.25em]
\begin{equation}
  \boldsymbol{\mathcal{L}}_\mathrm{smooth}=e^{|\nabla{\boldsymbol{\mathrm{I}}_\mathrm{ref}}|} \cdot \nabla{\boldsymbol{\mathrm{D}}_\mathrm{ref}} + e^{|\nabla^2{\boldsymbol{\mathrm{I}}_\mathrm{ref}}|} \cdot \nabla^2{\boldsymbol{\mathrm{D}}_\mathrm{ref}},
  \label{eq:geometric_consistency}
\end{equation}\\[-0.25em]
where $\boldsymbol{\mathrm{D}}_\mathrm{ref}$ means the depth map of $\boldsymbol{\mathrm{I}}_\mathrm{ref}$, and $\nabla, \nabla^2$ mean the first and second order derivatives.
\vspace{-1em}

\subsubsection{High-level Feature Alignment Consistency}
\label{sec:high-level}
We follow Huang \textit{et al.}\cite{huang2021m3vsnet} and utilize the high-level features alignment consistency to strengthen the robustness of our method. Containing more context information to show robust performance with various brightness, the high-level features are computed by the pre-trained networks, such as VGG-16\cite{simonyan2014very}, ResNet50\cite{he2016deep}, etc. Here VGG-16 is utilized to produce two-stage high-level features: features of 128 channels from the 8$^{th}$ layer, features of 256 channels from the 15$^{th}$ layer. According to the pixel-wise warping equation, as \mbox{Eq.~\ref{eq:photometric_pixel}}, the loss function of this module is:\\[-0.25em]
\begin{equation}
  \boldsymbol{\mathcal{L}}_{\mathrm{feature}}=\frac{1}{N}\sum\nolimits_{i=1}^{N}|\boldsymbol{F}_{\mathrm{ref}} - \hat{\boldsymbol{F}}_{\mathrm{src}}^{(i)}|\odot \boldsymbol{\mathrm{M}}_\mathrm{ref},
  \label{eq:feature_pixel}
\end{equation}\\[-0.25em]
where $\boldsymbol{F}_{\mathrm{ref}},\hat{\boldsymbol{F}}_{\mathrm{src}}^{(i)}$ mean the features of $\boldsymbol{\mathrm{I}}_\mathrm{ref}$ and the rendered features from $\boldsymbol{\mathrm{I}}_\mathrm{src}^{(i)}$.

\subsubsection{Overall Loss Formulation}
With the MVSNet architecture, the single-scale features and patches are used to construct the single-scale overall loss $\boldsymbol{\mathcal{L}}_\mathrm{SS}$:
\begin{equation}
  \begin{aligned}
    \boldsymbol{\mathcal{L}}_\mathrm{SS}  = & \lambda_1 \cdot \boldsymbol{\mathcal{L}}_\mathrm{patch} + \lambda_2 \cdot \boldsymbol{\mathcal{L}}_\mathrm{SSIM}  + \lambda_3 \cdot \boldsymbol{\mathcal{L}}_\mathrm{geometric} \\
                                            & +                                       \lambda_4 \cdot \boldsymbol{\mathcal{L}}_\mathrm{smooth}  +\lambda_5 \cdot \boldsymbol{\mathcal{L}}_\mathrm{feature},
    \label{eq:total_loss_single}
  \end{aligned}
\end{equation}
where $ \{\lambda_i\}_{i=1}^{5}$ mean the set of hyper-parameters of the single-scale overall loss.
\par
\noindent
With the multi-scale architecture, the multi-scale merged patches are integrated into the loss function and the independence of each stage in the compute graph is also broken so as there will be no interference for training between the different stages:
\begin{equation}
  \begin{aligned}
    \boldsymbol{\mathcal{L}}_\mathrm{MS} = \sum\nolimits_{k=1}^{K} \mu_k \cdot \boldsymbol{\mathcal{L}}_\mathrm{SS}^{(k)},
    \label{eq:total_loss_ms}
  \end{aligned}
\end{equation}
\par
\noindent
where $\mu_k$ means the hyper-parameter for the stage $k$, and $ \boldsymbol{\mathcal{L}}_\mathrm{SS}^{(k)}$ means the single-scale overall loss of the stage $k$.

\section{Experiments}
\label{sec:experiments}
\subsection{Datasets}
\par
\noindent
\textbf{DTU Multi-View Stereo Dataset}. DTU MVS datasets\cite{aanaes2016large} consist of indoor objects, containing 49 or 64 views with 7 different lighting conditions for each of the 124 scenes. The depth range is shorter compared to other datasets\cite{yao2020blendedmvs}, which is fixed from 425 $\mathrm{mm}$ to 935 $\mathrm{mm}$.
\par
\noindent
\textbf{Tanks and Temples Dataset}. Tanks and Temples\cite{Knapitsch2017} is a popular evaluation benchmark for image-based 3D reconstruction, which includes scenes from indoor rooms to outdoor environments and is split into intermediate and advanced.
\par
\noindent
\textbf{ETH3D Dataset}. ETH3D Stereo Benchmark\cite{schoeps2017cvpr} is a series of various indoor and outdoor scenes with both high and low resolutions. One character of this benchmark is the percentage of weakly-textured scenes is relatively high.

\begin{figure*}[t]
  \centering
  \begin{subfigure}{0.196\textwidth}
    \begin{minipage}[b]{1\textwidth}
      \centering
      \includegraphics[width=\linewidth]{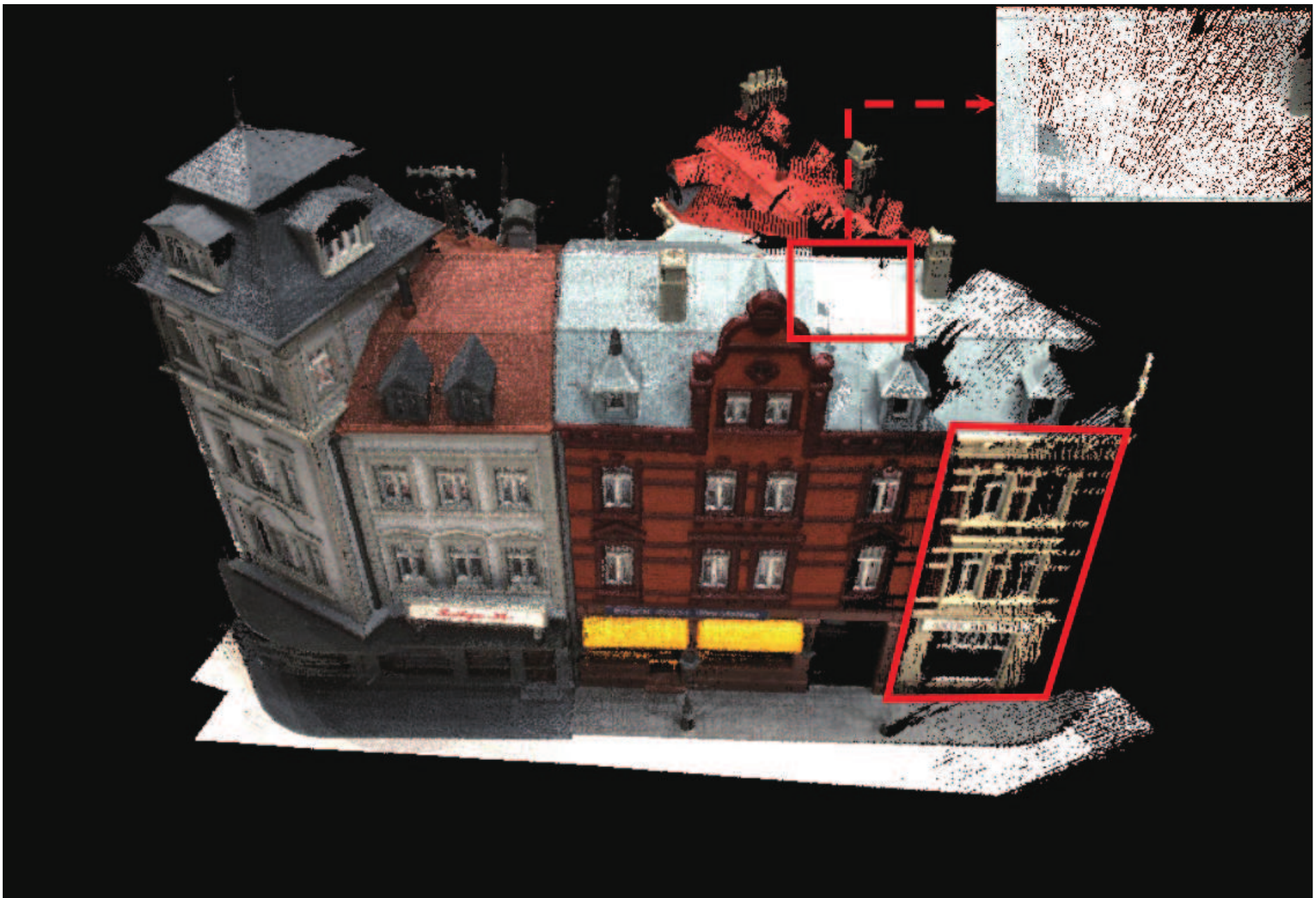}\\
      \includegraphics[width=\linewidth]{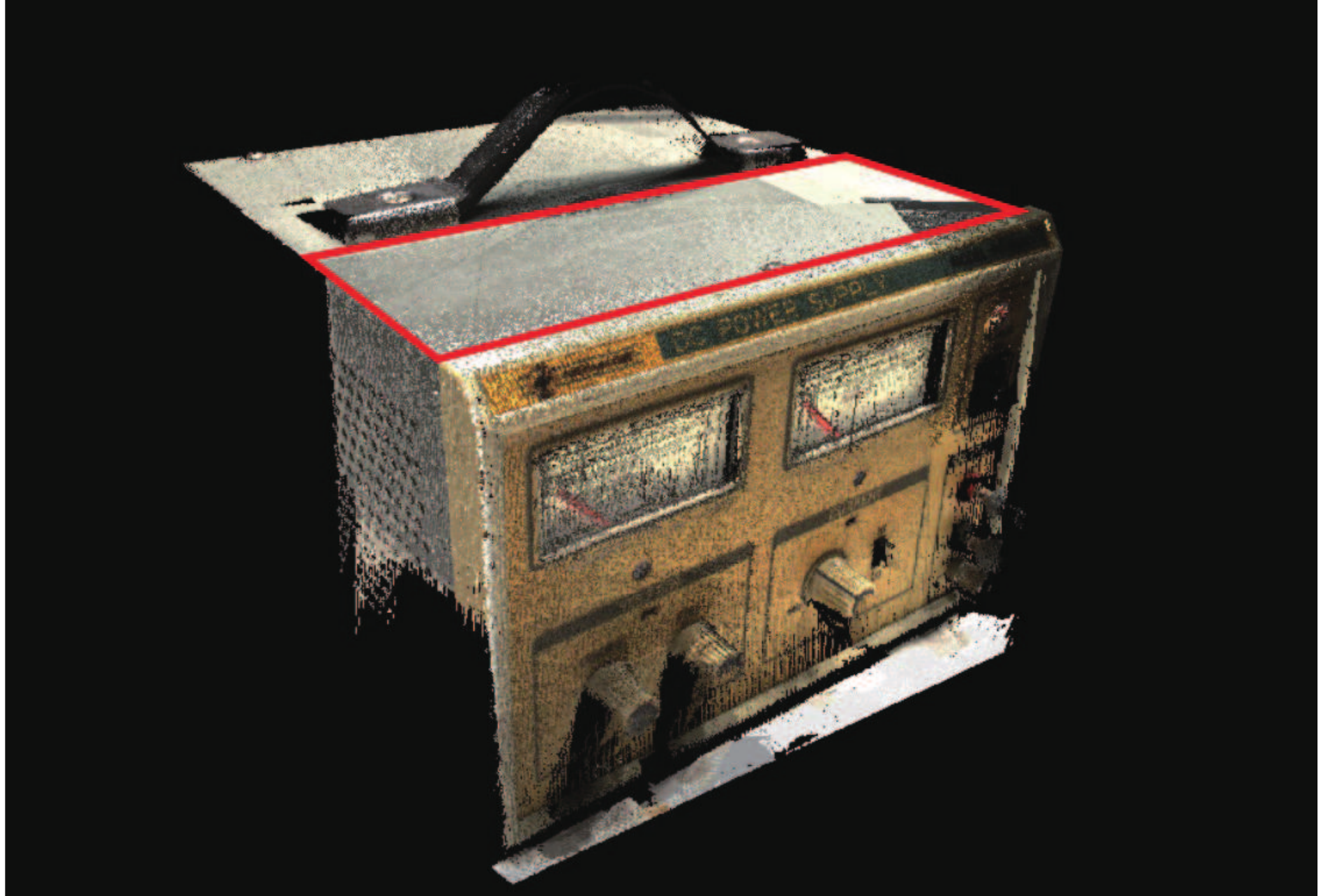}\\
      \includegraphics[width=\linewidth]{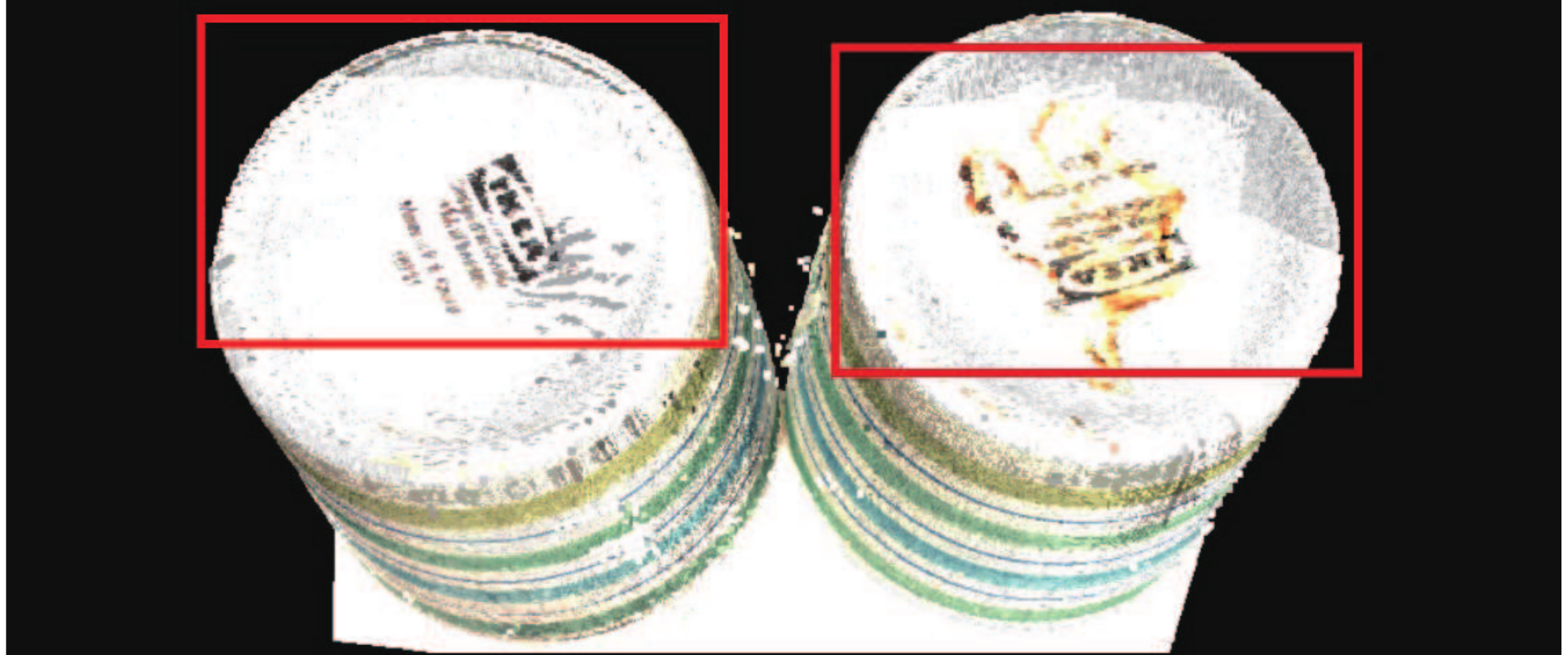}
    \end{minipage}
    \caption{Ground Truth}
  \end{subfigure}
  \hfill
  \begin{subfigure}{0.196\textwidth}
    \begin{minipage}[b]{1\textwidth}
      \centering
      \includegraphics[width=\linewidth]{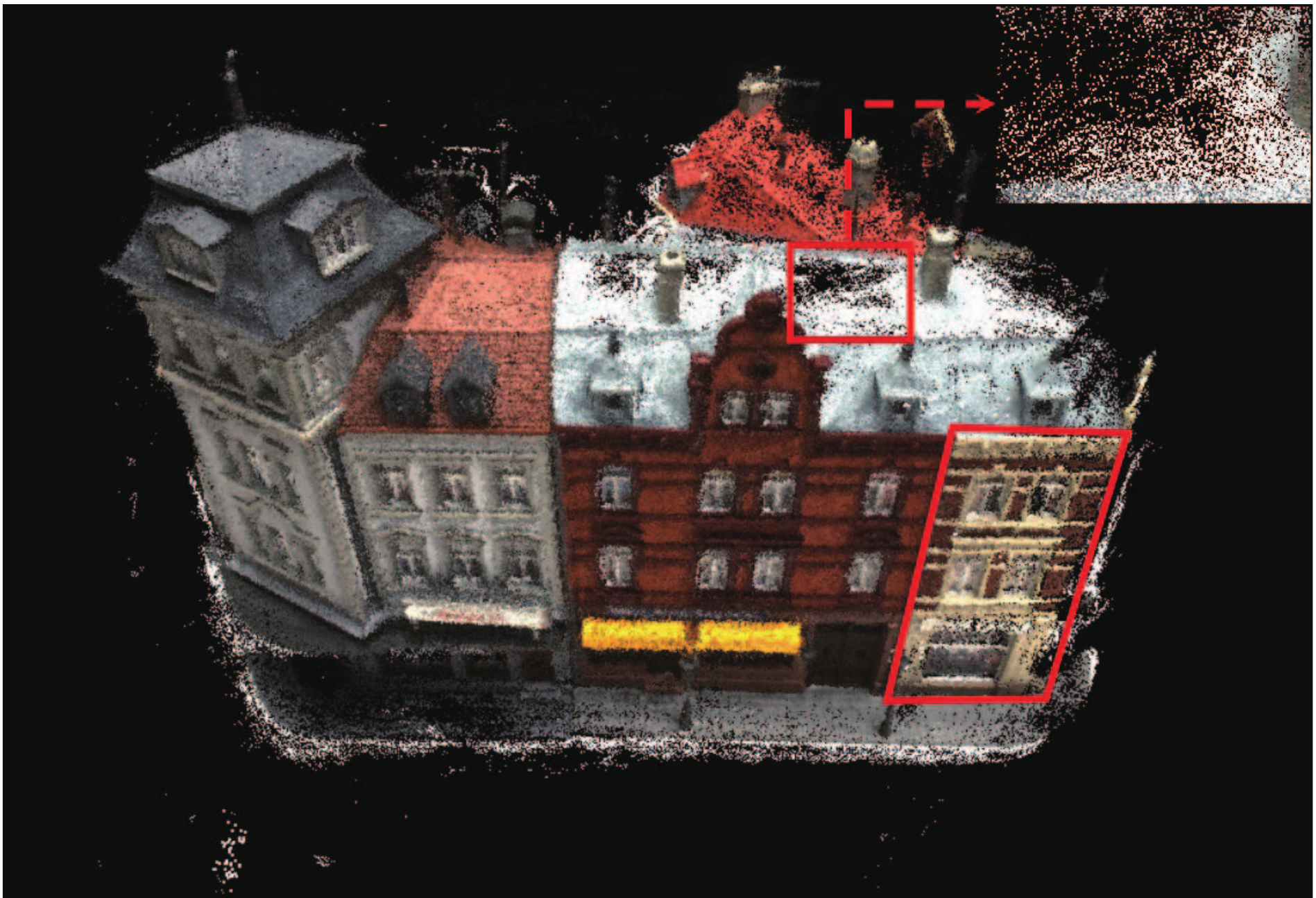}\\
      \includegraphics[width=\linewidth]{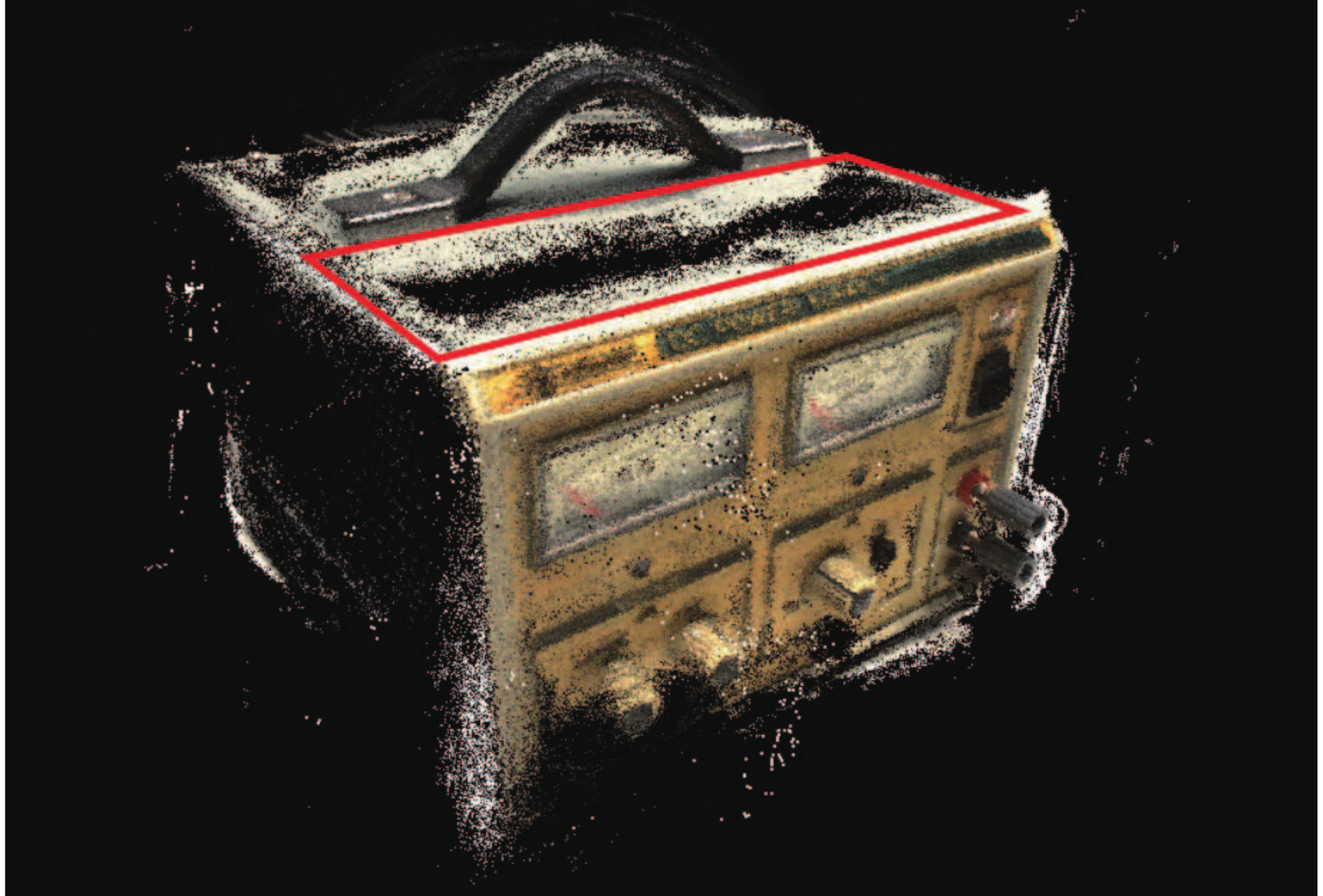}\\
      \includegraphics[width=\linewidth]{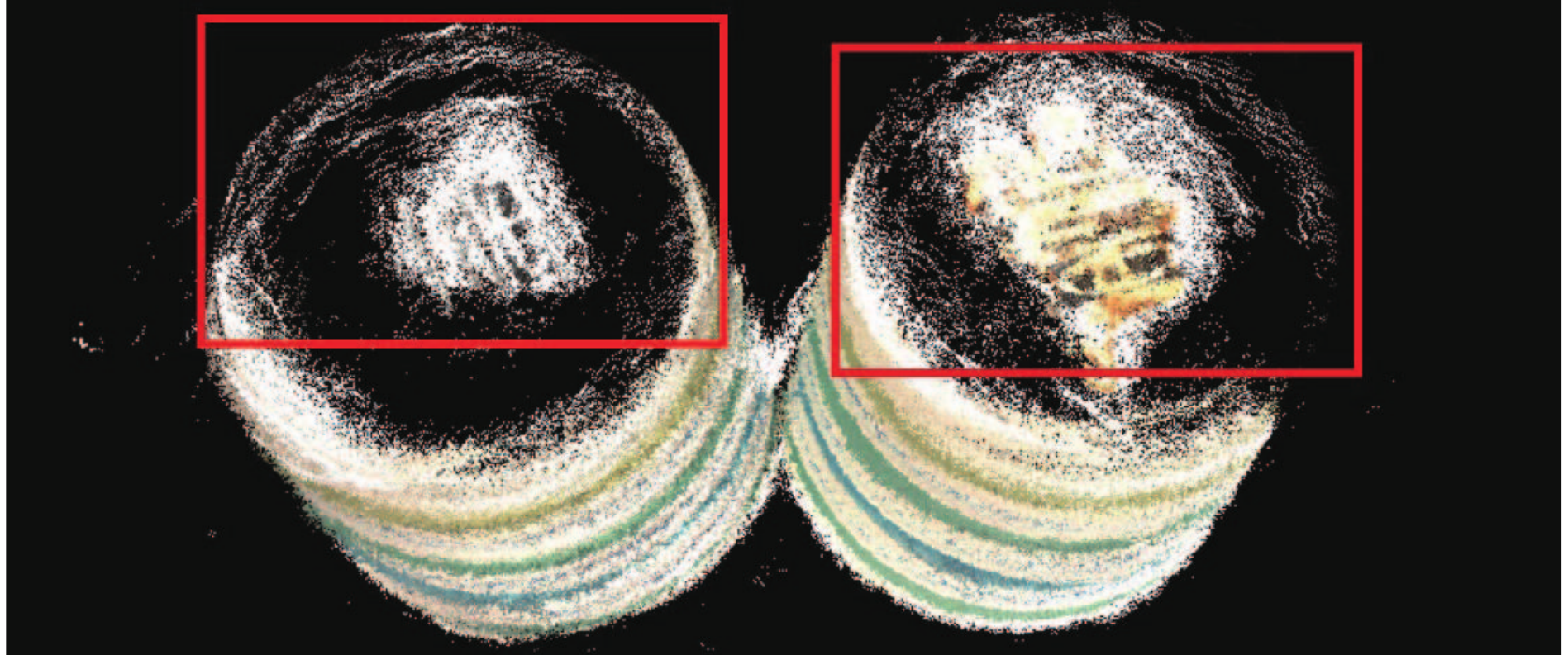}
    \end{minipage}
    \caption{M$^3$VSNet}
  \end{subfigure}
  \hfill
  \begin{subfigure}{0.196\textwidth}
    \begin{minipage}[b]{1\textwidth}
      \centering
      \includegraphics[width=\linewidth]{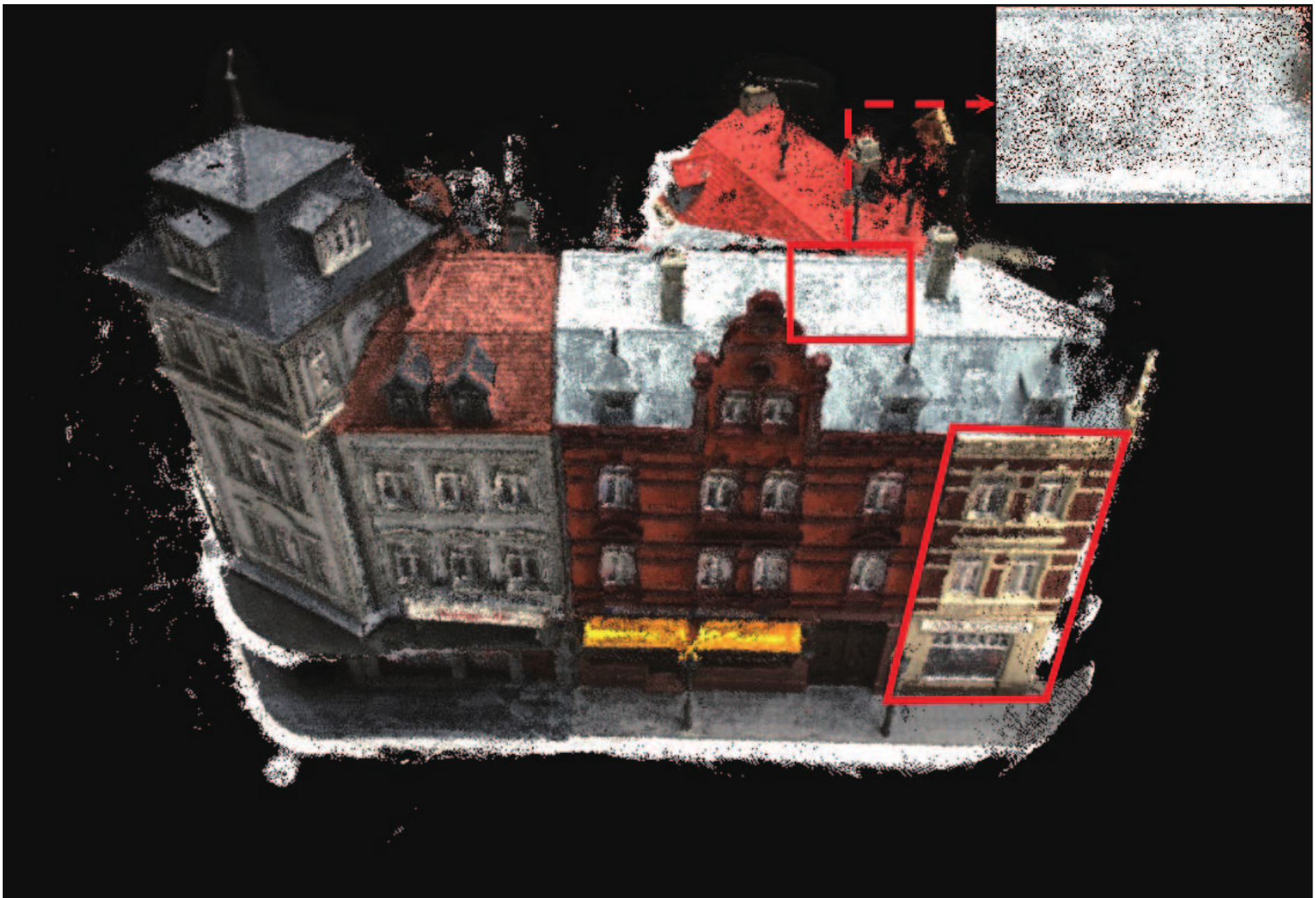}\\
      \includegraphics[width=\linewidth]{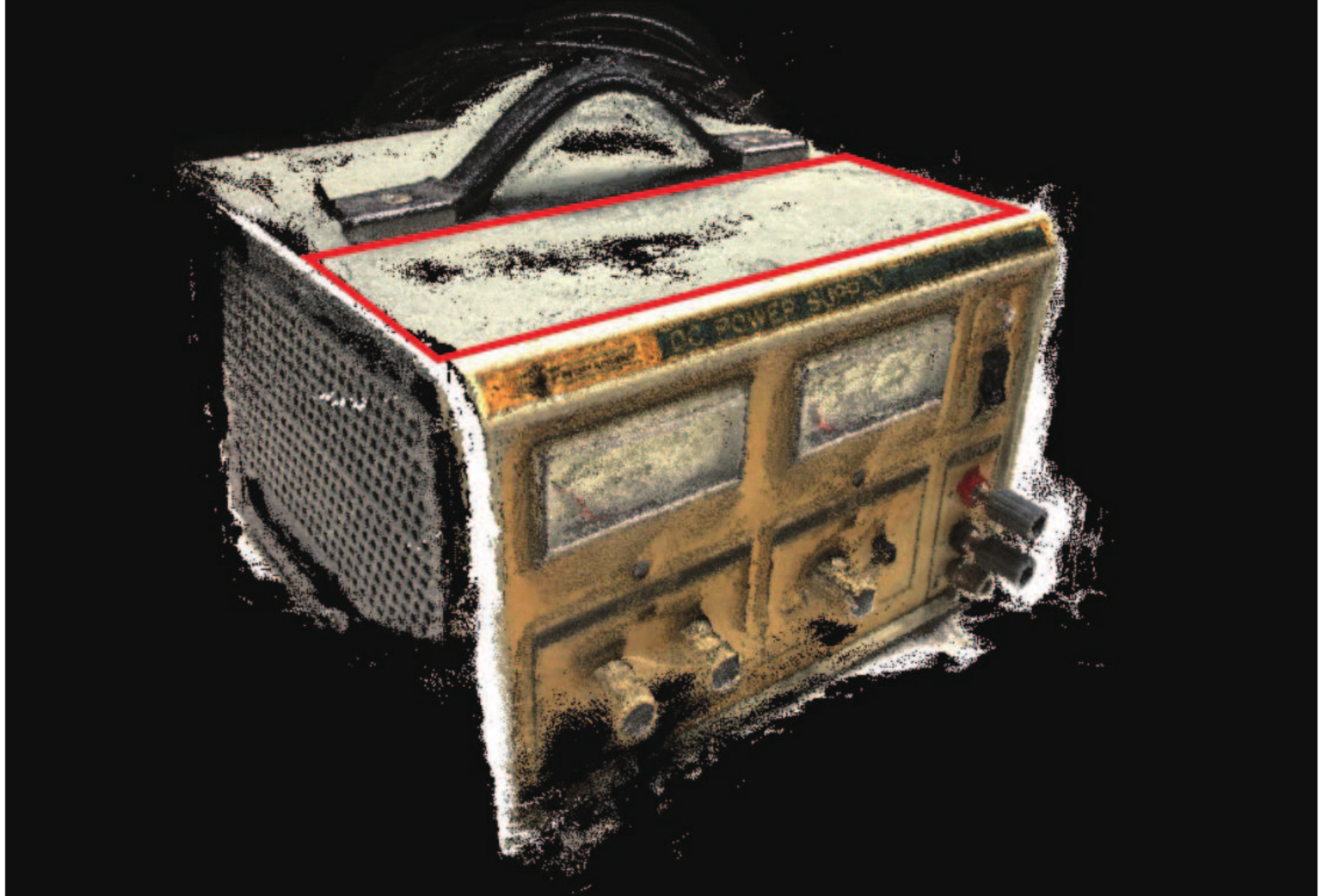}\\
      \includegraphics[width=\linewidth]{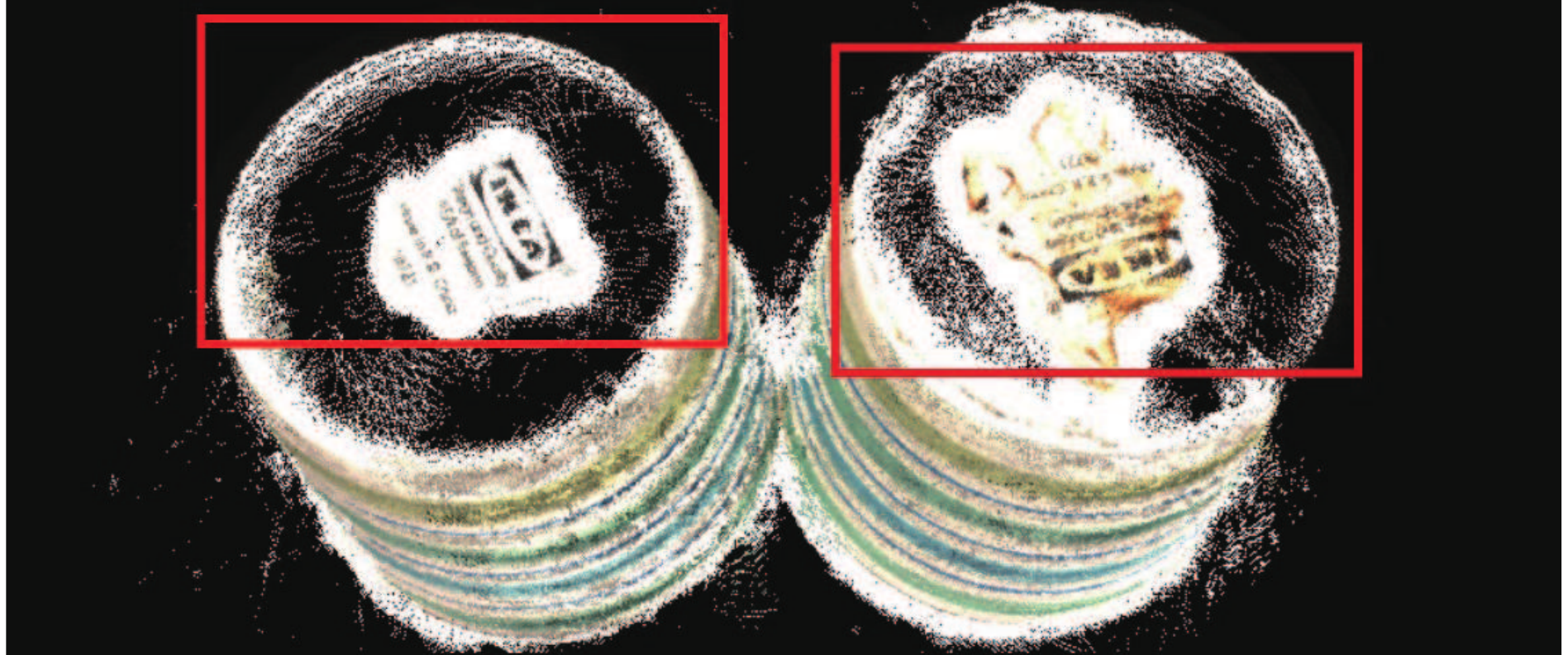}
    \end{minipage}
    \caption{\textit{PatchMVSNet (Ours)}}
  \end{subfigure}
  \hfill
  \begin{subfigure}{0.196\textwidth}
    \begin{minipage}[b]{1\textwidth}
      \centering
      \includegraphics[width=\linewidth]{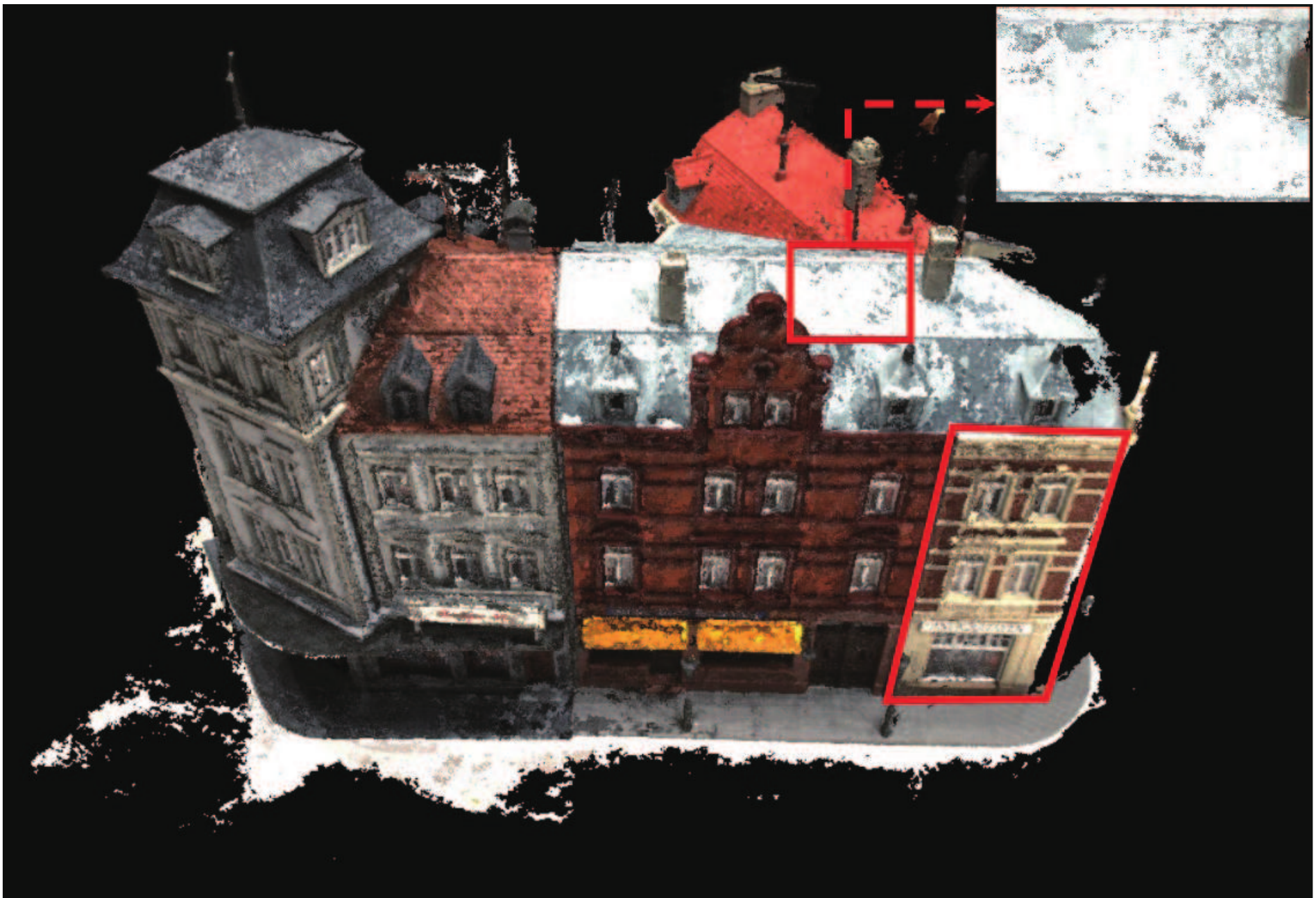}\\
      \includegraphics[width=\linewidth]{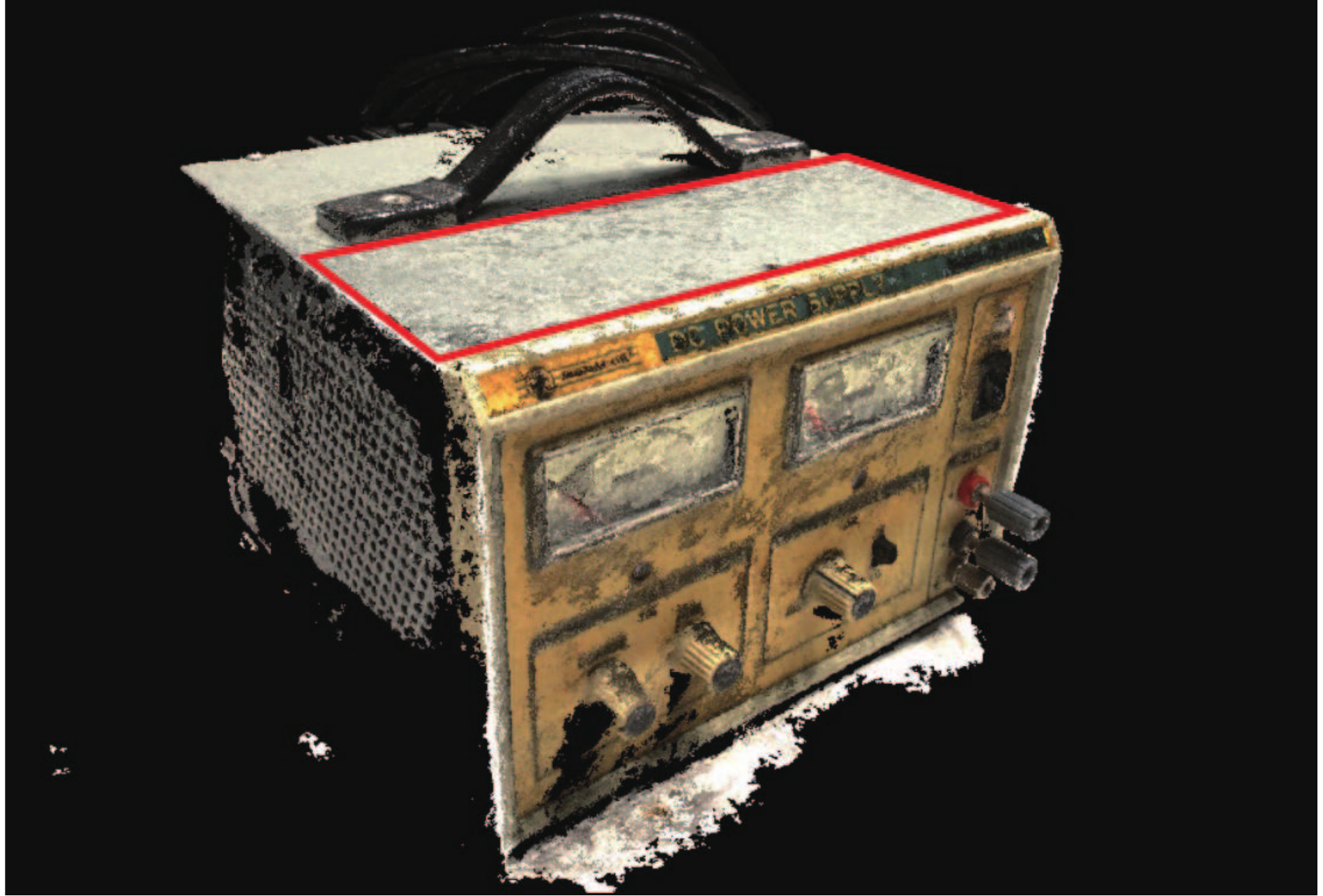}\\
      \includegraphics[width=\linewidth]{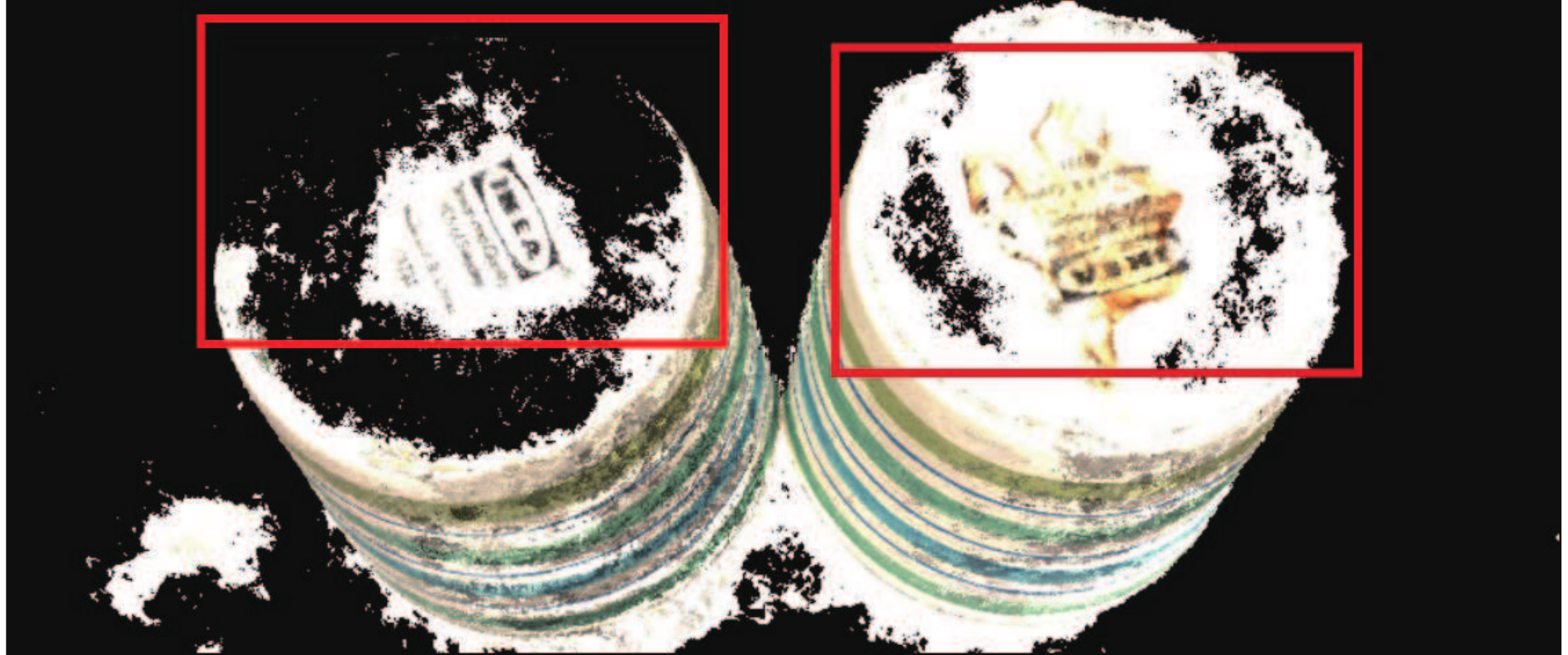}
    \end{minipage}
    \caption{CascadeMVSNet}
  \end{subfigure}
  \hfill
  \begin{subfigure}{0.196\textwidth}
    \begin{minipage}[b]{1\textwidth}
      \centering
      \includegraphics[width=\linewidth]{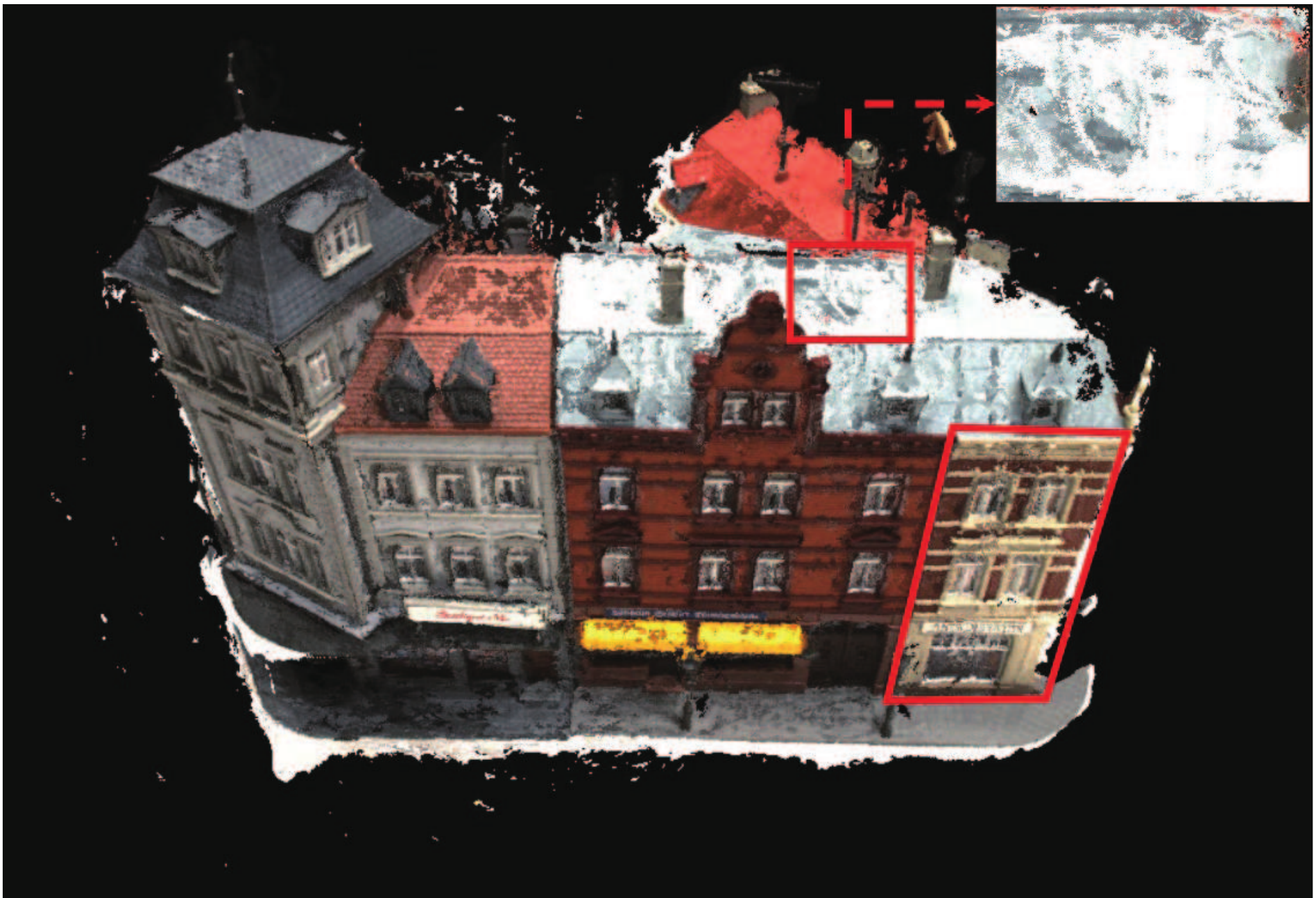}\\
      \includegraphics[width=\linewidth]{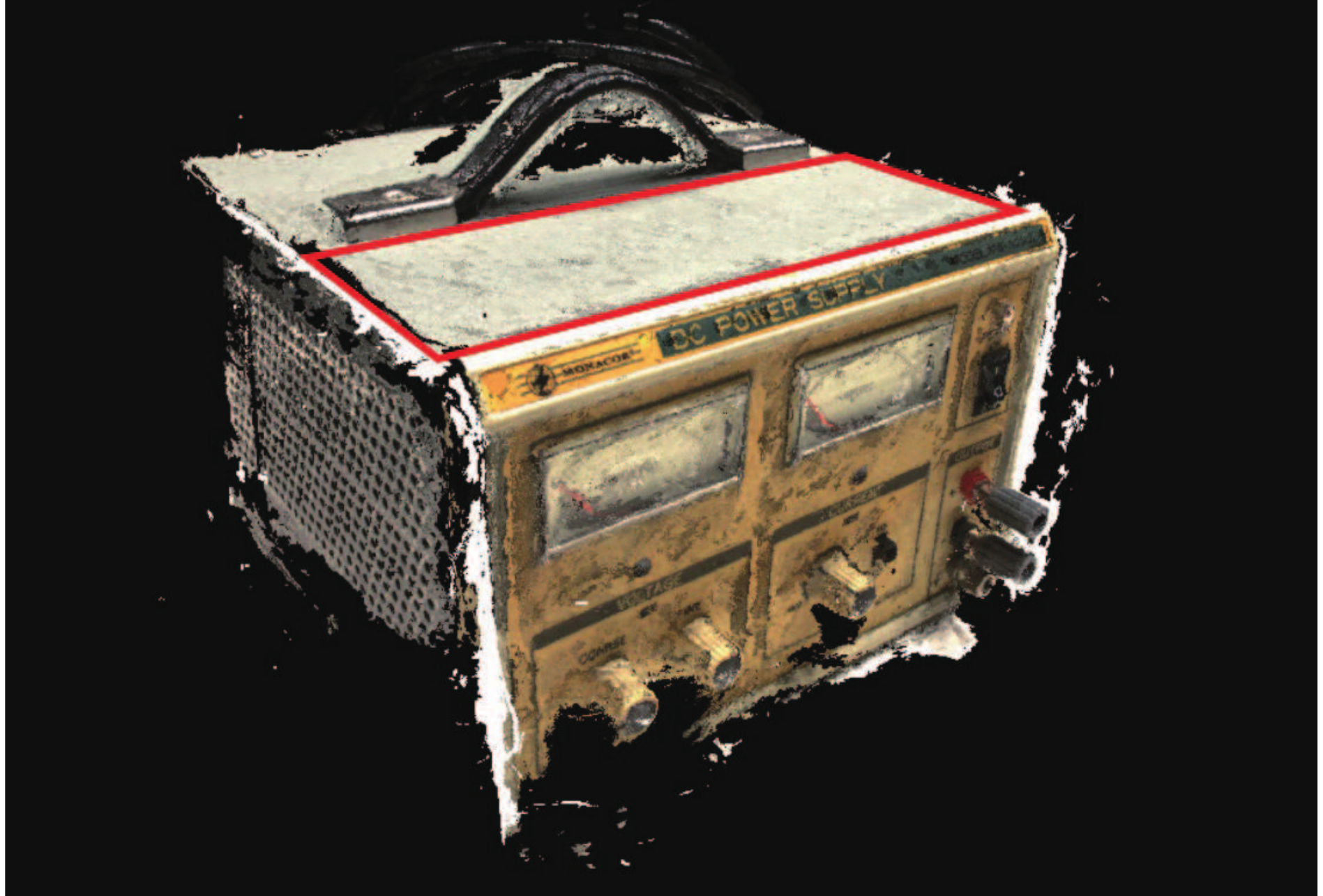}\\
      \includegraphics[width=\linewidth]{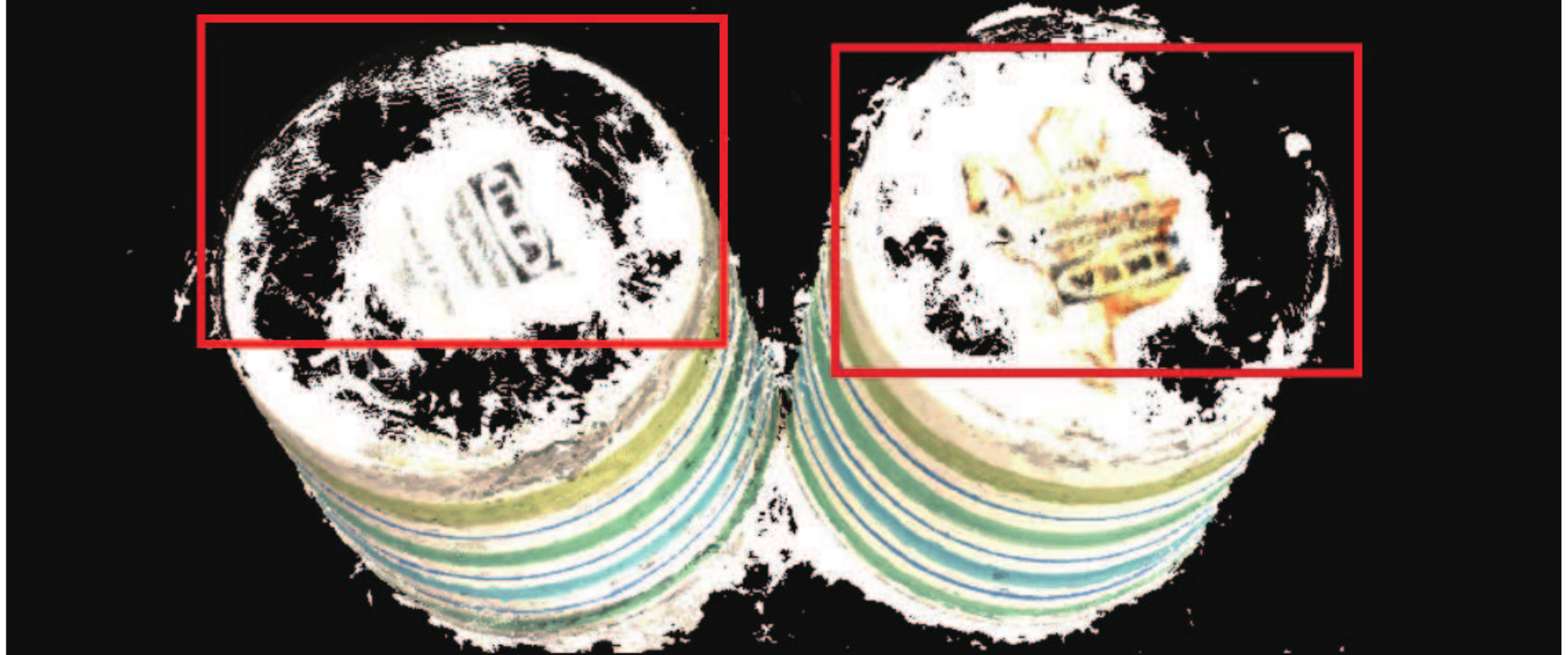}
    \end{minipage}
    \caption{\textit{PatchMVSNet-MS (Ours)}}
  \end{subfigure}\\[-0.75em]
  \caption{Visualization of the reconstructions comparison of the weakly-textured objects: (a) ground truth, (b) results of M$^3$VSNet\cite{huang2021m3vsnet} (unsupervised), (c) results of PatchMVSNet \textit{(Ours)} (unsupervised), (d) results of CascadeMVSNet\cite{gu2020cascade} (supervised), (e) results of PatchMVSNet-MS \textit{(Ours)} (unsupervised). Scan9, scan11 and scan48 (from the top to the bottom) in DTU evaluation datasets are selected, which include weakly-textured surfaces indicated in the red bounding boxes. In the top row, the upper red boxes indicate a highly reflecting surface in scan9 and our method can still reconstruct a relatively intact result, compared to the supervised method.}
  \label{fig:dtu-result}
  \vspace{-1em}
\end{figure*}

\subsection{Training on DTU}
\subsubsection{Implementation}
Our unsupervised strategy was implemented with the multi-scale architecture in this paper. We followed the setting of CascadeMVSNet\cite{gu2020cascade} and used a three-stage-scale structure. We separately sampled 48, 32 and 8 depth intervals and the scales of the depth interval are set to \mbox{4.24 $\mathrm{mm}$}, 2.12 $\mathrm{mm}$ and 1.06 $\mathrm{mm}$ from the first stage to the third stage and estimated a full-sized depth map of the reference image. The coefficients $\{\mu_{k}\}_{k=1}^{3}$ were set to 0.5, 1.0 and 2.0 from the first stage to the third stage. Considering the memory cost, we set the patch size to 3$\times$3. The hyper-parameters $\lambda_1,\lambda_2,\lambda_3,\lambda_4,\lambda_5$ were set to 0.8, 0.16, 1.0, 0.01, 1.0, respectively. For a fair comparison, we followed the splits of DTU datasets as MVSNet\cite{yao2018mvsnet} to train and evaluate our method. Our model was trained with 2 batches on 2 NVIDIA RTX 3090 cards for 6 epochs and the learning rate was set at 0.001 and was decreased by 50$\%$ for every two epochs.

\vspace{-1em}
\subsubsection{Performance on DTU Evaluation Datasets}
The evaluative metrics are \textit{accuracy} (the quality of the reconstructed 3D points), \textit{completeness} (the quantity of the captured surfaces on the object), and \textit{overall score} (the average value of accuracy and completeness).
\par
\noindent
\textbf{PatchMVSNet vs. Supervised and Traditional Methods.} In the quantitative comparisons, as illustrated in \mbox{Table~\ref{tab:compare_with_t_s}}, the completeness of our method outperforms all the traditional and supervised methods but PatchmatchNet\cite{wang2021patchmatchnet}, expressing the improved performance of our method in reconstructing more intact point clouds of various scenes. The overall score of our method is better than all the traditional methods and MVSNet\cite{yao2018mvsnet}. In the particular visualization of objects with the weak texture, as shown in \mbox{Figure~\ref{fig:dtu-result}}, our method can build the evenly-matched results, compared to the supervised multi-scale method\cite{gu2020cascade}. And our method can even get more fused points than it in scan9 and scan11 on DTU datasets.

\begin{table}[t]
  \centering
  \footnotesize
  \begin{tabular}{m{1.5cm}<{\centering}|m{3cm}<{\centering}|m{0.4cm}<{\centering} m{0.5cm}<{\centering} m{0.7cm}<{\centering}}
    \hline
                                 & Methods                                   & Acc.              & Comp.             & Overall           \\
    \hline
    \multirow{4}{*}{Traditional} & Furu\cite{furukawa2009accurate}           & 0.613             & 0.941             & 0.777             \\
                                 & Tola\cite{tola2012efficient}              & 0.342             & 1.190             & 0.766             \\
                                 & Camp\cite{campbell2008using}              & 0.835             & 0.554             & 0.694             \\
                                 & Gipuma\cite{galliani2015massively}        & \textbf{0.283}    & 0.873             & 0.578             \\
    \hline
    \multirow{6}{*}{Supervised}  & SurfaceNet\cite{ji2017surfacenet}         & 0.450             & 1.040             & 0.745             \\
                                 & MVSNet\cite{yao2018mvsnet}                & 0.396             & 0.527             & 0.462             \\
                                 & CIDER\cite{xu2020learning}                & 0.417             & 0.437             & 0.427             \\
                                 & CascadeMVSNet\cite{gu2020cascade}         & 0.325             & 0.385             & \underline{0.355} \\
                                 & PatchMatchNet\cite{wang2021patchmatchnet} & 0.427             & \textbf{0.277}    & \textbf{0.352}    \\
                                 & CVP-MVSNet\cite{yang2020cost}             & \underline{0.296} & 0.406             & \textbf{0.352}    \\
    \hline
    Unsupervised                 & PatchMVSNet-MS \textit{(Ours)}            & 0.538             & \underline{0.365} & 0.451             \\
    \hline
  \end{tabular}\\[-0.75em]
  \caption{Quantitative comparisons with traditional and supervised methods on DTU evaluation datasets. The \textbf{bold} number means the best in the single column, and the \underline{underline} indicates the second-best.}
  \label{tab:compare_with_t_s}
\end{table}
\begin{table}[t]
  \centering
  \footnotesize
  \begin{tabular}{c|c|ccc}
    \hline
     & Methods                                & Acc.           & Comp.          & Overall        \\
    \hline
    \multirowcell{5}{End-to-                                                                     \\End}  & Khot\cite{khot2019learning}                     & 0.881          & 1.073          & 0.997          \\
     & MVS$^{2}$\cite{dai2019mvs2}            & 0.760          & 0.515          & 0.637          \\
     & M$^3$VSNet\cite{huang2021m3vsnet}      & 0.636          & 0.531          & 0.583          \\
     & JDACS \cite{xu2021self}                & 0.571          & 0.515          & 0.543          \\
     & PatchMVSNet-MS \textit{(Ours)}         & \textbf{0.538} & \textbf{0.365} & \textbf{0.451} \\
    \hline
    \multirowcell{4}{Multi-                                                                      \\Stage} & Meta\cite{mallick2020learning}                  & 0.594          & 0.779          & 0.687          \\
     & Yang\cite{Yang_2021_CVPR}              & 0.308          & 0.418          & 0.363          \\
     & Xu\cite{xu2021digging} + MVSNet        & 0.470          & 0.430          & 0.450          \\
     & Xu\cite{xu2021digging} + CascadeMVSNet & \textbf{0.354} & \textbf{0.354} & \textbf{0.354} \\
    \hline
  \end{tabular}\\[-1em]
  \caption{Quantitative comparisons with existing unsupervised methods on DTU evaluation datasets. The \textbf{bold} number means the best in the column of each class.}
  \label{tab:compare_with_us}
  \vspace{-2.25em}
\end{table}
\par
\noindent
\textbf{PatchMVSNet vs. Other Unsupervised Methods.} Existing unsupervised methods can be categorized into two classes. The first class uses the end-to-end strategy, which focuses on building an end-to-end learning framework. The second class leverages the multi-stage strategy, which trains the model with an additional post-processing stage, leveraging the assistants from the third-party inputs, such as the ground truth\cite{mallick2020learning}, a rendered mesh\cite{Yang_2021_CVPR} or optical flow\cite{xu2021digging}. In the quantitative comparisons, as listed in \mbox{Table~\ref{tab:compare_with_us}}, our method outperforms all the end-to-end methods both in \textit{accuracy} and \textit{completeness}. Compared to the multi-stage methods, our method still has an advantage of \textit{completeness}, which outperforms all methods but the best and is only a little worse than it (3.1\%).
\par
\noindent
Compared to M$^3$VSNet\cite{huang2021m3vsnet} and implemented with the MVSNet architecture, our method could build a more precise and complete point cloud than it (see \mbox{Figure~\ref{fig:dtu-result}}), especially in those weakly-textured surfaces. Implemented with the multi-scale architecture, our method could preserve an even more complete point cloud than the implementation with the single-scale architecture.
\vspace{-1em}

\subsubsection{Generalization}
The generalization of our proposed PatchMVSNet was firstly trained on DTU training datasets and then evaluated on \textit{Tanks and Temples} and \textit{ETH3D Stereo Benchmark} without any fine-tune.
\par
\begin{figure}[t]
  \centering
  \begin{subfigure}{0.205\textwidth}
    \centering
    \includegraphics[height=3.3cm]{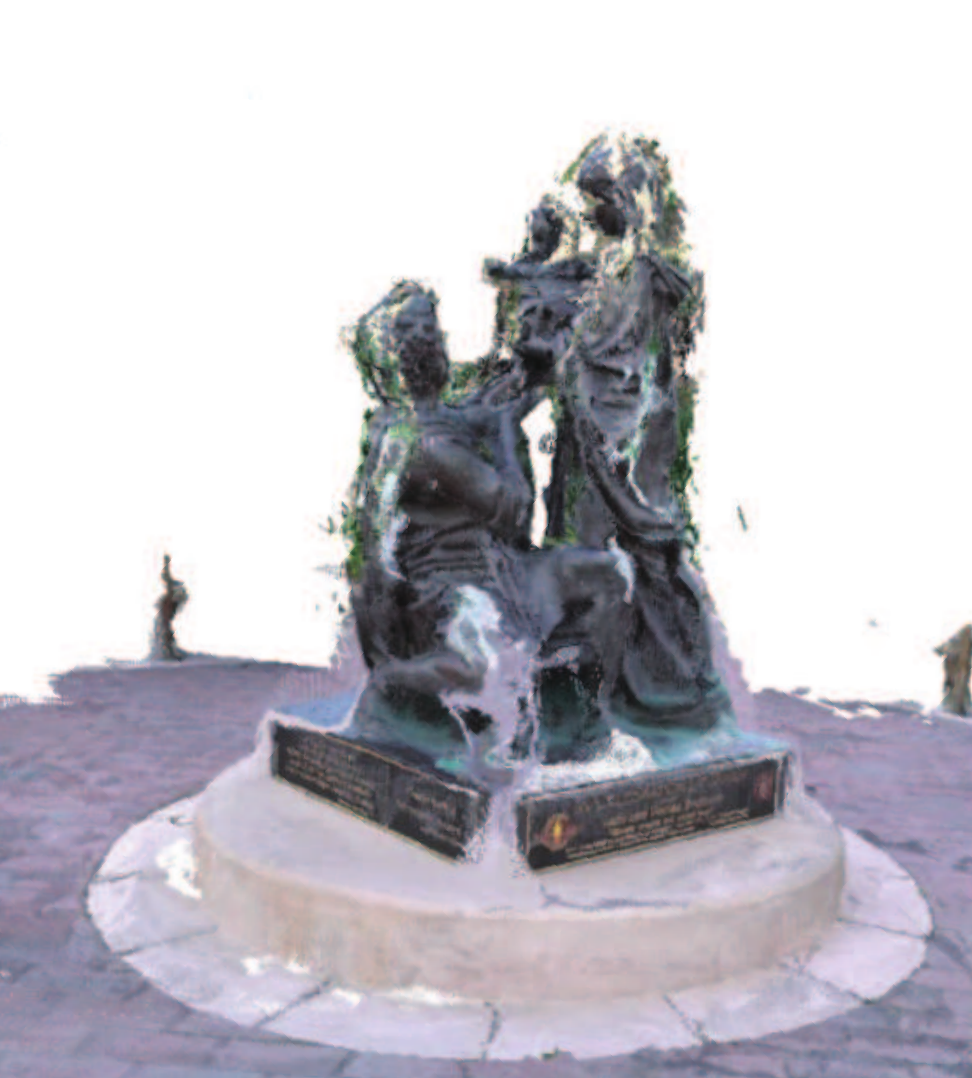}
    \caption{Family}
  \end{subfigure}
  \hfill
  \begin{subfigure}{0.265\textwidth}
    \centering
    \includegraphics[height=3.3cm]{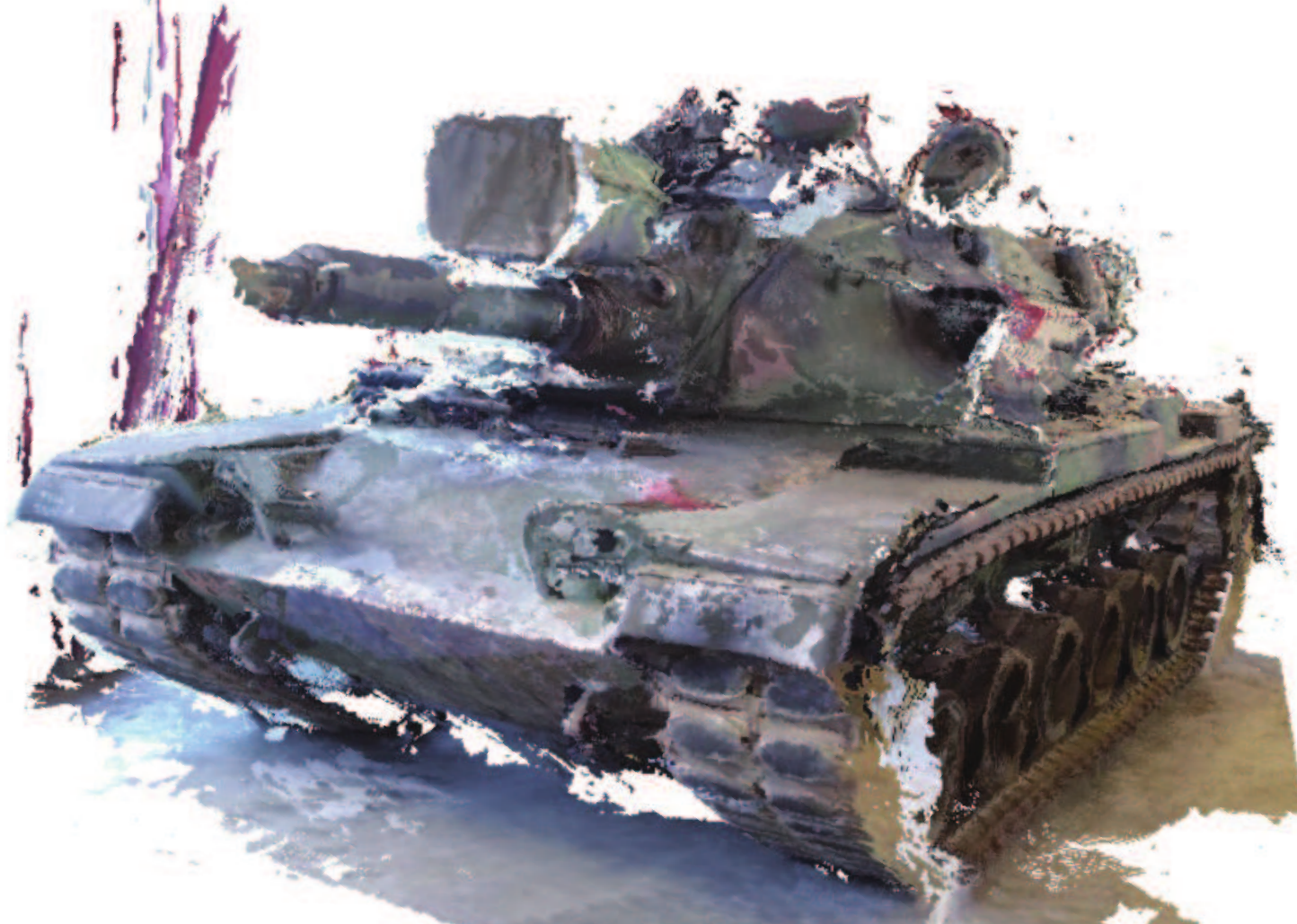}
    \caption{M60}
  \end{subfigure}
  \begin{subfigure}{0.47\textwidth}
    \centering
    \includegraphics[width=1\linewidth]{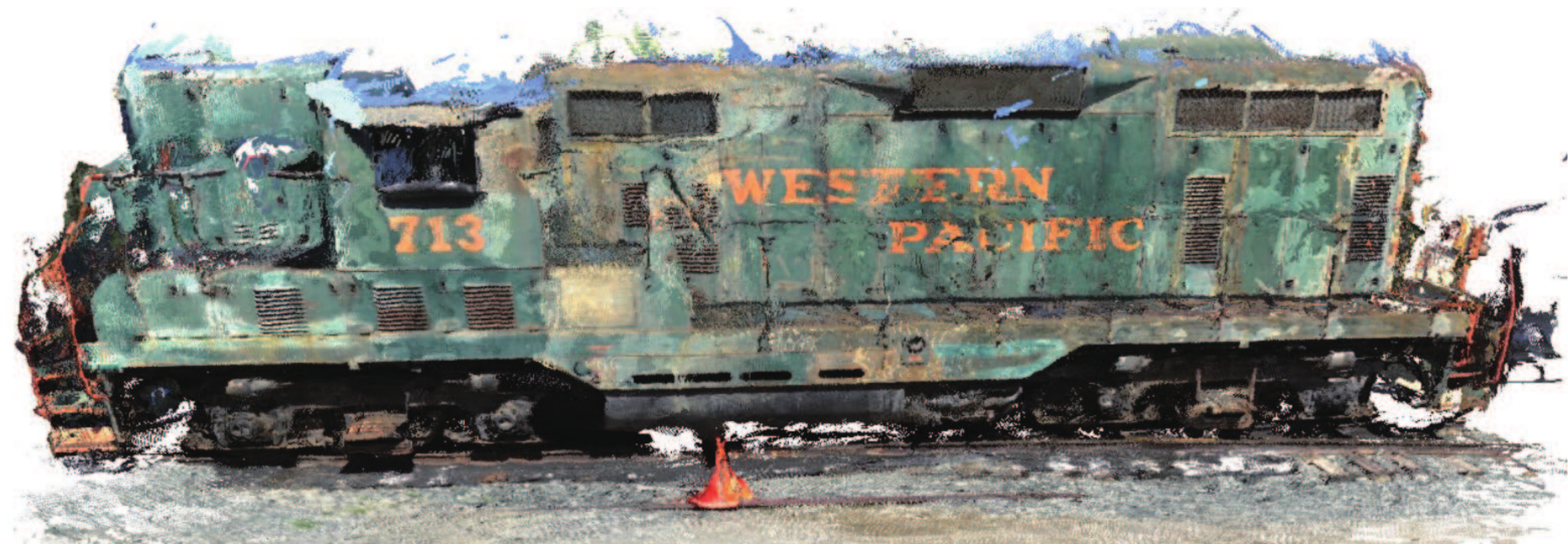}
    \caption{Train}
  \end{subfigure}\\[-0.5em]
  \caption{Visualization of \textit{Tanks and Temples - intermediate group}: (a) Family, (b) M60, (c) Train.}
  \label{fig:tat-ours}
  \vspace{-1.5em}
\end{figure}

\noindent
\textbf{Tanks and Temples}. The quantitative results are shown in the \mbox{Table~\ref{tab:tat}}, and our method has the better performance than MVS$^2$ and M$^3$VSNet, where the reconstructions of \textit{Family} (improved by 16.6\%), \textit{Francis} (improved by 48.04\%) and \textit{Horse} (improved by 57.8\%) are more intact than the remaining methods. The visualization of the point clouds of the partial results of \textit{Tanks and Temples - intermediate group} are also shown in \mbox{Figure~\ref{fig:tat-ours}}.
\begin{table}[t]
  \footnotesize
  \centering
  \begin{tabular}{c|m{0.6cm}<{\centering}m{0.7cm}<{\centering}m{0.7cm}<{\centering}m{0.6cm}<{\centering}m{0.6cm}<{\centering}}
    \hline
    Methods                           & Mean           & Family         & Francis        & Horse          & Train          \\
    \hline
    MVS$^2$\cite{dai2019mvs2}         & 37.21          & 47.74          & 21.55          & 19.50          & 29.72          \\
    \hline
    M$^3$VSNet\cite{huang2021m3vsnet} & 37.67          & 47.74          & 24.38          & 18.76          & 30.31          \\
    \hline
    PatchMVSNet \textit(Ours)         & \textbf{40.26} & \textbf{55.66} & \textbf{33.87} & \textbf{30.19} & \textbf{35.15} \\
    \hline
  \end{tabular}\\[-1em]
  \caption{Part of the quantitative comparisons on \textit{Tanks and Temples - intermediate group}. The \textbf{bold} number indicates the best of each column. Full result will be listed in the supplementary material.}
  \label{tab:tat}
  \vspace{-1.25em}
\end{table}


\par
\begin{figure}[t]
  \centering
  \begin{subfigure}{0.45\textwidth}
    \centering
    \includegraphics[width=1\linewidth]{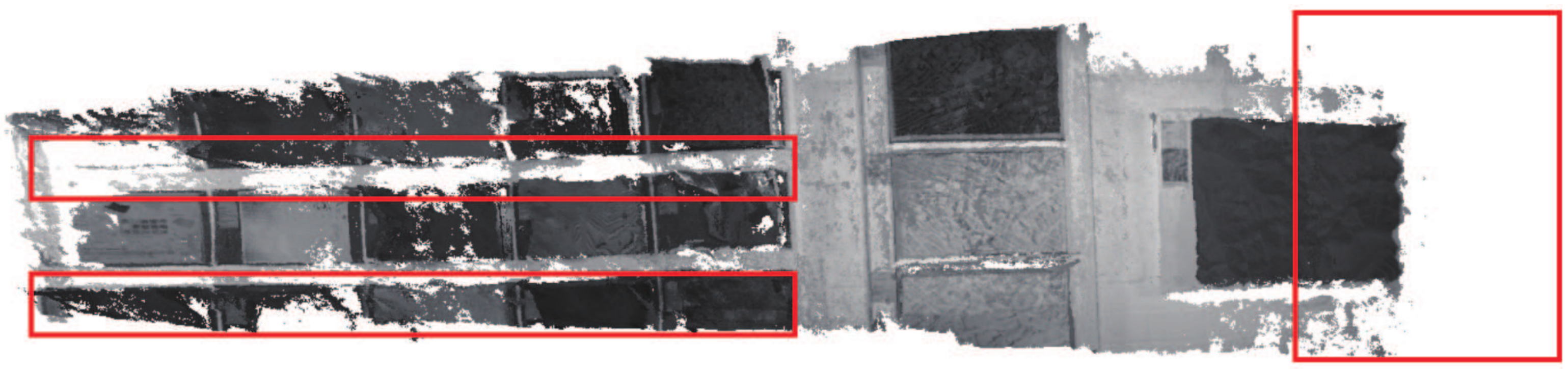}
    \caption{COLMAP + OpenMVS}
  \end{subfigure}
  \begin{subfigure}{0.45\textwidth}
    \centering
    \includegraphics[width=1\linewidth]{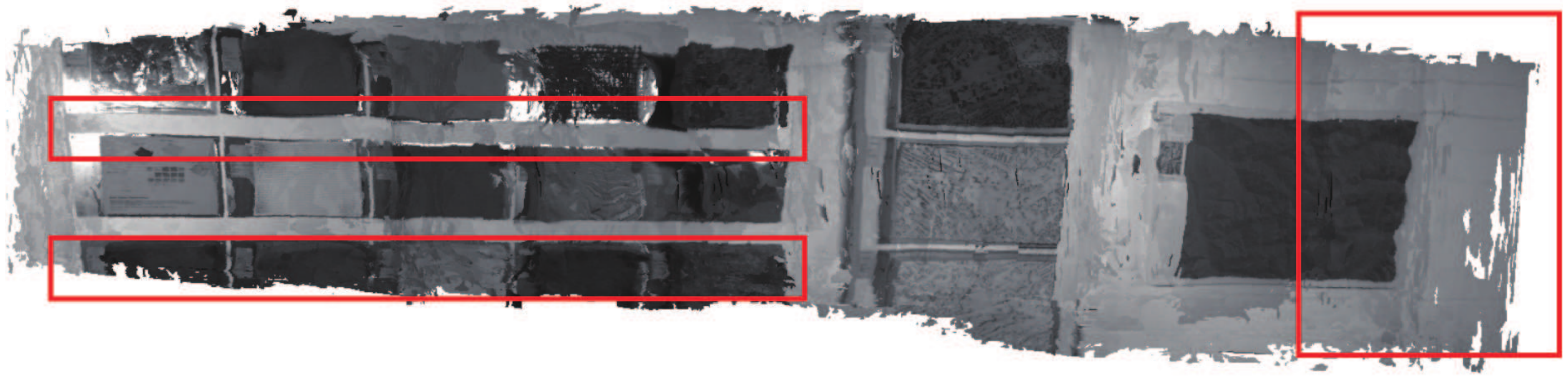}
    \caption{PatchMVSNet-MS \textit{(Ours)}}
  \end{subfigure}\\[-0.5em]
  \caption{Visualization of \textit{terrains}: (a) the result of COLMAP\cite{schoenberger2016mvs} + OpenMVS\cite{openmvs2020}, (b) the result of our method. As shown in red boxes, our method can build an intact result even for weak texture.}
  \label{fig:terr}
\end{figure}

\noindent
\textbf{ETH3D Stereo Benchmark}. Our method shows a powerful performance on this benchmark, as it contains lots of weakly-textured scenes, like the building facades and the indoor walls. As shown in \mbox{Table~\ref{tab:eth3d_low}}, our method achieves the best performance of published unsupervised methods at \textit{ETH3D low-resolution-many-view benchmark} until November 15$^{th}$, 2021, which ranks 28$^{th}$ after the submission of CascadeMVSNet\cite{gu2020cascade} (27$^{th}$). Compared to M$^3$VSNet\cite{huang2021m3vsnet}, our method outperforms it by the mean score (improved by 213.8\%). In the visualization of the fused point clouds, our method can successfully reconstruct the weakly-textured surfaces of terrains, which is better than OpenMVS\cite{openmvs2020} in the visual effect (see \mbox{Figure~\ref{fig:terr}}).
\begin{table}[t]
  \vspace{0.5em}
  \footnotesize
  \centering
  \begin{tabular}{c|cccccc}
    \hline
    Methods                           & Mean              & deli. & electro & terrains \\
    \hline
    DPSNet\cite{im2019dpsnet}         & 39.62             & 23.74 & 29.07   & 70.20    \\
    CascadeMVSNet\cite{gu2020cascade} & \textbf{67.59}    & 59.55 & 73.85   & 79.22    \\
    \hline
    M$^3$VSNet\cite{huang2021m3vsnet} & 18.17             & 7.77  & 15.59   & 62.02    \\
    PatchMVSNet-MS \textit{(Ours)}    & \underline{57.02} & 51.47 & 56.11   & 74.12    \\
    \hline
  \end{tabular}\\[-0.5em]
  \caption{Part of quantitative comparisons on \textit{ETH3D low-resolution-many-view benchmark}. The upper two methods are supervised methods, while the lower two are unsupervised methods. The \textbf{bold} number means the best in all listed methods, and the \underline{underline} means the second-best. Full result will be shown in the supplementary material.}
  \label{tab:eth3d_low}
  \vspace{-1em}
\end{table}

\subsection{Ablation Study}
The Ablation study is conducted on DTU evaluation datasets to inspect the effectiveness of our loss components and the influence of different numbers of the selected source images.
\vspace{-0.5em}
\subsubsection{Effectiveness of the Loss Components}
The baseline of our method was implemented with CascadeMVSNet, trained by the pixel-wise photometric consistency and the high-level feature alignment, and then we added the depth smooth loss and the robust geometric consistency loss to show the effectiveness of geometric constraints. Lastly, we replaced the pixel-wise loss with the patch-wise loss to show the improvement from the patches. We trained each of these combinations for 5 epochs, as listed in \mbox{Table ~\ref{tab:ablation-dtu}}, all combinations reached the best performance, compared to all end-to-end unsupervised methods in \mbox{Table ~\ref{tab:compare_with_us}}, which showed the effectiveness of our novel loss modules. Compared to the baseline, adding the geometric consistency loss can increase the completeness of reconstructions but slightly impair the accuracy, since the depth map is not reliably estimated from the pixel-wise loss. After we replaced the pixel-wise loss with the patch-wise loss (the size of the patch is 3$\times$3), both accuracy and completeness are improved, proving the robustness of the patch-wise photometric consistency.
\begin{table}[t]
  \footnotesize
  \centering
  \begin{tabular}{c|m{0.5cm}<{\centering}m{0.6cm}<{\centering}m{0.7cm}<{\centering}}
    \hline
    Combinations                                                                        & Acc.           & Comp.          & Overall        \\
    \hline
    Baseline(Pixel)                                                                     & 0.594          & 0.430          & 0.512          \\
    \hline
    Baseline(Pixel) + $\mathcal{L}_\mathrm{smooth}$ + $\mathcal{L}_\mathrm{geometric}$  & 0.614          & 0.419          & 0.517          \\
    \hline
    Baseline(Patch) + $\mathcal{L}_\mathrm{smooth}$  + $\mathcal{L}_\mathrm{geometric}$ & \textbf{0.538} & \textbf{0.365} & \textbf{0.451} \\
    \hline
  \end{tabular}\\[-0.75em]
  \caption{Quantitative comparisons on DTU evaluation datasets with different combinations of loss modules. All experiments share the same setting, which used 4 views for the depth estimation and 4 views for the 3D point reservation. The \textbf{bold} number means the best.}
  \label{tab:ablation-dtu}
  \vspace{-3em}
\end{table}

\vspace{-0.5em}
\subsubsection{View Numbers}
We trained our model with 4 views, while in the evaluation phase, we changed the view numbers in the depth estimation. For the threshold of the view-consistency, we also set different view numbers to preserve the 3D points. The quantitative comparisons with different view numbers are listed in \mbox{Table~\ref{tab:ablation-view-points}}. Increasing the number of the views for depth estimation can improve the accuracy since it provides more redundant information of visibility. While decreasing the threshold of the view-consistency can increase the completeness, where our implementation that used 4 views for depth estimation and 3 views for points preserving reaches the best completeness score in all published unsupervised methods on DTU evaluation datasets. For seeking a compromise, we eventually chose the strategy of the second row as our official evaluation result on DTU evaluation datasets.
\begin{table}[t]
  \footnotesize
  \centering
  \begin{tabular}{c|ccc}
    \hline
    View Number        & Acc.  & Comp. & Overall        \\
    \hline
    $N_1 =$5, $N_2 =$4 & 0.514 & 0.397 & 0.456          \\
    \hline
    $N_1 =$4, $N_2 =$4 & 0.538 & 0.365 & \textbf{0.451} \\
    \hline
    $N_1 =$4, $N_2 =$3 & 0.631 & 0.330 & 0.481          \\
    \hline
  \end{tabular}\\[-0.5em]
  \caption{Quantitative comparisons on DTU evaluation datasets with different view numbers. $N_1$ means the number of views selected for depth estimation, and $N_2$ means the number of views selected for points preserving. The \textbf{bold} number means the best.}
  \label{tab:ablation-view-points}
  \vspace{-1.5em}
\end{table}

\vspace{-0.5em}
\section{Conclusion}
This paper aims to the reconstruction on weakly-textured surfaces by unsupervised MVS and proposes \mbox{PatchMVSNet} with novel loss functions. First, a patch-wise photometric consistency loss is used to infer a robust depth map of the reference image. Then the robust cross-view geometric consistency is utilized to further decrease the matching ambiguity. Moreover, the high-level feature alignment is leveraged to alleviate the uncertainty of the matching correspondences. For the implementations without loss of generality, MVSNet and CascadeMVSNet are selected as our backbones. Trained on DTU datasets and tested on popular datasets, such as \textit{DTU evaluation datasets, Tanks and Temples}, our method reaches a comparable performance to the state-of-the-art methods and has an advantage in completeness. Experiments on \textit{ETH3D Stereo Benchmark} show that our method has the powerful performance among unsupervised methods for large-scale weakly-textured reconstructions. In the future work, our attention will be paid to the strategy of further merging the pixel and the patches of multiple sizes, which will utilize more precise information from the images for the high-quality reconstruction.

\newpage
{\small
  \bibliographystyle{ms}
  \bibliography{ms}
}

\end{document}